\documentclass[twoside]{article}
\pdfoutput=1
\usepackage[accepted]{aistats2012}
\usepackage{amsmath,amssymb,float,multirow} 
\usepackage[final]{graphicx} 
\usepackage[numbers,sort&compress,comma]{natbib} 
\usepackage{pdfpages}
\usepackage{wrapfig}
\usepackage[tight]{subfigure} 
\usepackage{verbatim} 
\usepackage{epsfig}
\usepackage{algorithm}
\usepackage[noend]{algorithmic}
\usepackage{floatflt}
\usepackage{paralist}

\newcommand{\matC}{\ensuremath{\mathcal{C}}}

\newcommand{\dpar}{\ensuremath{(d)}}

\newcommand{\wCountD}[1]{\ensuremath{\sharp_{#1}^{\dpar}}}

\newcommand{\piD}[1]{\ensuremath{\pi_{#1}^{\dpar}}}
\newcommand{\pivD}{\ensuremath{\boldsymbol{\pi}^{\dpar}}}
\newcommand{\pivv}[2]{\ensuremath{\boldsymbol{\pi}_{#1}^{#2}}}

\newcommand{\fD}[1]{\ensuremath{\hat{f}_{#1}^{\dpar}}}
\newcommand{\fvD}{\ensuremath{\mathbf{\hat{f}}^{\dpar}}}

\newcommand{\fvv}[2]{\ensuremath{\mathbf{\hat{f}}_{#1}^{#2}}}

\newcommand{\fnotD}[1]{\ensuremath{\mathbf{\hat{f}}_{-#1}^{(d)}}}
\newcommand{\fvnotD}{\ensuremath{\mathbf{\hat{f}}^{(-d)}}}
\newcommand{\fvnotDji}{\ensuremath{\mathbf{\hat{f}}_{ji}^{(-d)}}}

\newcommand{\cD}[1]{\ensuremath{\hat{c}_{#1}^{\dpar}}}
\newcommand{\cvD}{\ensuremath{\mathbf{\hat{c}}^{\dpar}}}
\newcommand{\cnotD}[1]{\ensuremath{\mathbf{\hat{c}}_{-#1}^{\dpar}}}
\newcommand{\cv}[1]{\ensuremath{\mathbf{\hat{c}}^{(#1)}}}
\newcommand{\cvv}[2]{\ensuremath{\mathbf{\hat{c}}_{#1}^{#2}}}

\newcommand{\wvD}{\ensuremath{\mathbf{w}^{\dpar}}}
\newcommand{\wnD}{\ensuremath{w_n^{\dpar}}}

\newcommand{\znD}{\ensuremath{z_n^{\dpar}}}
\newcommand{\zvD}{\ensuremath{\mathbf{z}^{\dpar}}}

\newcommand{\pinotD}[1]{\ensuremath{\boldsymbol{\pi}_{-#1}^{\dpar}}}
\newcommand{\nD}[1]{\ensuremath{n_{#1}^{\dpar}}}

\newcommand{\remove}[1]{}

\begin{document}

%

%

\twocolumn[

\aistatstitle{Concept Modeling with Superwords}

\aistatsauthor{ Khalid El-Arini \And Emily B. Fox \And Carlos Guestrin}

\aistatsaddress{Computer Science Department \\ Carnegie Mellon
  University\And Department of Statistics \\ The Wharton School \\ University of
  Pennsylvania \And Machine Learning Department \\ Carnegie Mellon University} ]

\begin{abstract}  
In information retrieval, a fundamental goal is to transform a
document into concepts that are representative of its content.  
The term ``representative'' is in itself challenging to define, and
various tasks require different granularities of concepts.
  In this paper, we aim to model concepts that are sparse over the vocabulary, and that flexibly adapt their
  content based on other relevant \emph{semantic} information such as
  textual structure or associated image features.  We explore a
  Bayesian nonparametric model based on nested beta processes that
  allows for inferring an unknown number of strictly sparse concepts.
  The resulting model provides an inherently different representation
  of concepts than standard LDA (or HDP) based topic models, and
  allows for direct incorporation of semantic features.  We
  demonstrate the utility of this representation on multilingual blog
  data and the Congressional Record.
\end{abstract}

\section{Introduction}

Information overload is a ubiquitous problem that affects nearly all members of society, from researchers
sifting through millions of scientific articles to Web users trying to gauge public opinion by reading blogs.
Even as information retrieval (IR) methods evolve to move beyond the traditional Web search paradigm to more varied
retrieval tasks focused on combating this overload, they remain reliant on suitable document representations.
While all representations ultimately distill the contents of a document collection into fundamental \emph{concepts},
representing the atomic units of information, a particular choice of representation 
can have potentially drastic consequences on performance.

For instance, a common desideratum of retrieval tasks is \emph{diversity} in the result set~\cite{MMR}.  Here, 
the document representation must be expressive enough such that it is recognizable when
two documents are about the same idea.  A fine-grained representation (e.g., individual words or named entities)
may lead to many concepts with the same meaning.  For example, ``President Obama,'' ``Barack Obama,'' ``Mr. President''
and ``POTUS'' are all distinct named entities that refer to the 44th President of the United States, but may end up as distinct
concepts depending on the representation.  On the other end of the spectrum, coarse-grained representations, such as
topics from a topic model~\cite{blei:lafferty:bookchapter}, may conflate together many ideas that are only vaguely related.  This 
vagueness is particularly
a problem for systems that attempt to \emph{personalize} results to
user's individual tastes, and as such need to
estimate a user's level of interest in each concept (e.g.,~\cite{AmrPersonalization,TDN}).

In this paper, we seek to \emph{learn} concept representations at an appropriate level of granularity, 
representing each concept as a set of words that are functionally equivalent for the particular task at hand. We take
a cue from the computer vision community~\cite{superpixels} and refer to such concepts as \emph{superwords}.  
We desire the following characteristics from our model:
\begin{compactenum}
\item The number of potential ideas that are to be modeled as concepts is unknown and unbounded, and thus our
model should be able to handle this uncertainty. 
\item Our model should specify a probabilistic interpretation for how much each document is about any given concept,
allowing for seamless incorporation into IR systems.
\item We should be able to easily encode semantic information about the vocabulary into our model, based on
the idea that two words occurring in the same concept should share a meaning in some underlying semantic space.
\item The same concept can be represented differently in different documents (i.e., it can contain slightly
different sets of words).
\end{compactenum}

Our approach addresses these properties by relying on the machinery of Bayesian nonparametric methods.  In particular,
we use a \emph{nested beta process} prior to provide strict sparsity in the set of concepts used in a document and 
the set of words associated with each concept, while at the same time allowing for uncertainty in the number of 
concepts~\cite{Hjort:90,Thibaux:07,GriffithsGhahramani:05,Rodriguez:08,Jordan:10}.  Such a prior encourages sharing
of concepts and word choices among documents, but provides flexibility for documents to differ.  For instance, 
Democrats may say ``healthcare reform'' while Republicans may opt to
say ``Obamacare,'' but both are referring to the same concept.  Previous models assume that 
topics are the same for each document, and so to model such a phenomenon,
they either create multiple topics to refer to the same idea, or else
a single conflated topic with probability mass on words informed by
both populations.  We avoid both undesirable options by using a more expressive prior.

Moreover, as described above, there is inherent uncertainty in the granularity of the concepts.  
Specifically, does one choose more concepts, each with fewer words, or fewer concepts, each with more words?  
This is a question of \emph{identifiability} in the nested beta process.  Thus, to further inform our sought-after sparsity 
structure, we augment the prior with a \emph{semantic feature matrix} in which each word in the vocabulary has an 
associated observed feature vector.  This matrix has the additional benefit of allowing us to fuse multiple 
sources of information.  For example, the feature vector may capture sentence co-occurrence of words in the vocabulary.  Such information 
harnesses the structure of the text lost in the simple bag-of-words formulation.  Alternatively (or additionally), this feature vector can 
include non-textual information, such as features of images of each word, or features learned from user feedback that can be
useful for personalization. The semantic feature matrix is modeled as a weighted combination of latent 
\emph{concept semantic features}, where the weighting is based on word assignments to each concept across the corpus,
thereby encoding semantic similarity into concept membership.

\begin{figure}[t]
\begin{center}
\subfigure[\label{fig:obamacareNBP}]{\includegraphics[width=2.9cm]{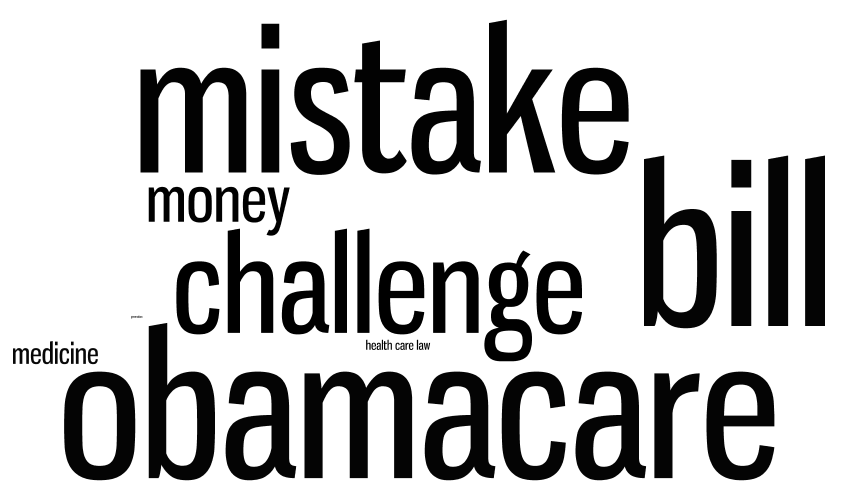}}
\subfigure[\label{fig:obamacareHDP}]{\includegraphics[width=3.3cm]{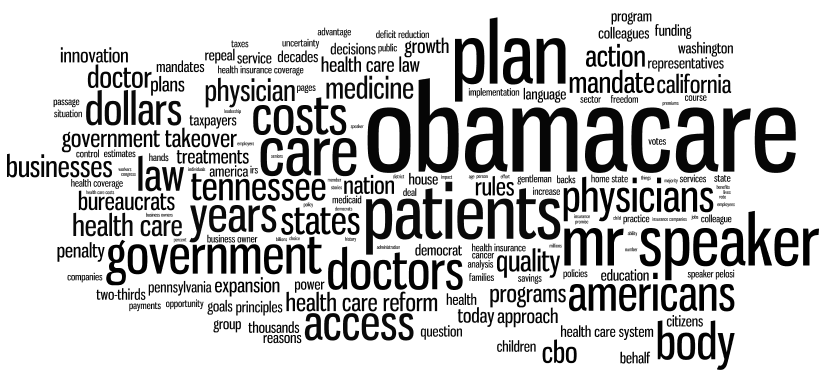}}\\
\subfigure[\label{fig:obamacaredNBP-study}]{\includegraphics[width=3.3cm]{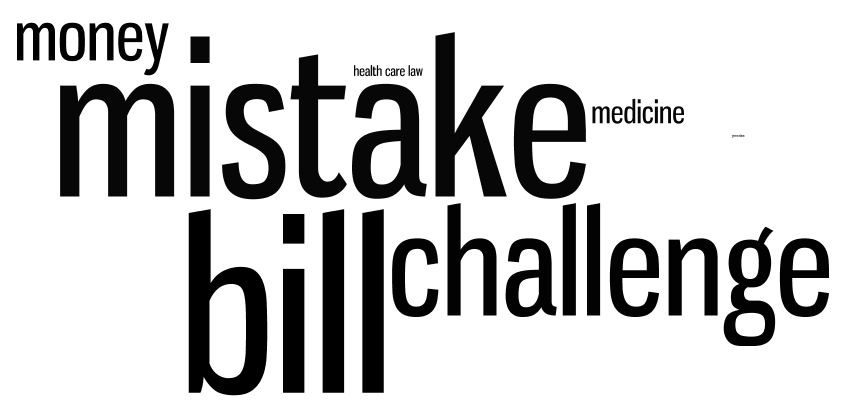}}
\subfigure[\label{fig:obamacareHDP-study}]{\includegraphics[width=2.7cm]{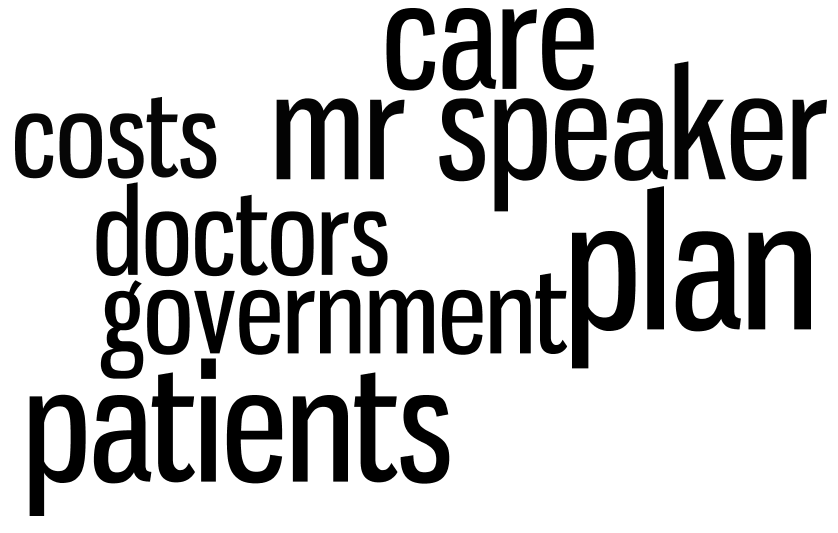}}
\vspace*{-5mm}
\caption{\footnotesize{The top row contains word clouds comparing 
a concept learned using our model (a) with a topic from an
HDP (b), both involving the word ``Obamacare.''
The bottom row has the same concept and topic, but as displayed to
users in our user study, described in Sec.~\ref{sec:congress}.\label{fig:hdp-nbp-compare}}}
\vspace*{-5mm}
\end{center}
\end{figure}

The model we propose in this paper is fundamentally different from topic models and other generative models of text in that
we represent each concept as a \emph{sparse set of words}, rather than as topics that are distributions over an
entire vocabulary. As such, our concepts can be more coherent and
focused than traditional topics. For example,
the top row of Figure~\ref{fig:hdp-nbp-compare} compares
 a concept learned via our
model with a corresponding topic from a hierarchical Dirichlet process
(HDP)~\cite{Teh:06}, both learned on the same Congressional Record corpus
(cf. Sec.~\ref{sec:congress}).\footnote{Throughout this paper, word clouds are used to illustrate
topics and concepts, where the size of a word is proportional to its weight or prevalence
in the topic or concept.} We see in (a) that our model is able to focus
on the salient idea of the word ``Obamacare,'' in that it is a
pejorative term used by Republicans to describe President Obama's
health care reform package.  The HDP topic in (b) provides a much vaguer and
more diffuse representation.  Additionally, following 
Williamson et al.~\cite{FTM}, our prior allows us to decouple the prevalence of a concept from its strength.  In other
words, in our model, it is possible for a rare concept to be highly important to the documents that contain it,
which is a characteristic difficult to obtain in traditional topic modeling approaches.

To recap, the main contributions of this paper are:
\begin{compactitem}
\item A novel use of a nested beta process prior to define concepts
in a documents as sparse sets of words;
\item The ability to elegantly incorporate semantic side information
to guide concept formation;
\item An MCMC inference procedure for learning concepts from data; and
\item Empirical results--including a user study--on both 
multilingual blog data and the Congressional Record, showing
the efficacy of our approach.
\end{compactitem}
\section{Nested Beta Processes} \label{sec:nestedBP} 
We wish to model the situation where there are an unknown (and
unbounded) collection of concepts in the world, and each document is about
some sparse subset of them.  A natural way to model this uncertainty
and unboundedness in a probabilistic manner is to look to Bayesian
nonparametric methods.  
For instance, Dirichlet processes (DPs) have long been used as priors
for mixture models in which the number of mixture components is
unknown~\cite{Antoniak:74}. However, in our case, rather than assigning each
observation to a single cluster as done in DPs, we require a 
\emph{featural} model, where each document can be made up of several
concepts.  As such, we base our model on a nested version of the \emph{beta process}~\cite{Hjort:90},
described in this section.

\subsection{Beta Process - Bernoulli Process}
We consider the situation where there is a countably infinite number of
concepts in the world, each represented by a coin with a particular
bias, $\omega_j$, and a set of attributes that define the concept, $\boldsymbol{\psi}_j$.
A process for assigning concepts to a particular document could thus be to flip each of the coins, and if
coin $j$ lands heads, we assign concept $j$ to the document.  This
process of flipping coins to assign concepts to each document is known
formally as the \emph{Bernoulli process}, since we have a $\hat{c}_j^{(d)} \sim Bernoulli(\omega_j)$
draw for each concept, where $\hat{c}_j^{(d)}$ indicates that concept
$j$ is on for document $d$.  

As we do not know the values of the coin biases $\omega_j$ and concept
attributes $\boldsymbol{\psi}_j$, we wish to place a prior over them that
encodes our desire for a sparse set of active concepts per
document.  Specifically,  if we let,
\begin{equation}
B = \sum_{j=1}^\infty \omega_j\delta_{\boldsymbol{\psi}_j},
\end{equation}
we exploit of the fact that the Bernoulli
process has a conjugate prior known as the beta process, and write $B
\sim BP(b, B_0)$ to indicate that $B$ is distributed according to a
beta process with \emph{concentration parameter} $b > 0$ and a
base measure $B_0$ over some measurable space $\boldsymbol{\Psi}$.  
By construction, the biases $\omega_j$ lie in the interval (0,1), and
thus if the mass of the base measure, $\alpha_\omega =
B_0(\boldsymbol{\Psi})$, is finite, then $B$ has finite expected
measure and we obtain our desired concept sparsity.\footnote{It is important
to note that, unlike a Dirichlet process base measure,
$B_0(\boldsymbol{\Psi})$ need not be equal to 1.  The beta
process is a \emph{completely random measure}~\cite{Kin1967}, where realizations on
disjoint sets are independent random variables.}

Formally, the beta process is defined as a realization of a
nonhomogenous Poisson process with rate measure defined as the
product of the base measure $B_0$ and an improper beta distribution.
In the special case where the base measure
contains discrete atoms $i$, with associated measure $\lambda_i$,
then a sample $B \sim BP(b, B_0)$ necessarily contains the atom, with
associated weight $\omega_i \sim Beta(b \lambda_i, b(1 - \lambda_i))$ (cf.~\cite{Thibaux:07}).

\subsection{Indian Buffet Process}
A Bernoulli process realization $\mathbf{\hat{c}}^{(d)}$ from our prior
determines the subset of concepts that are active for document
$d$.  As in Thibaux and Jordan~\cite{Thibaux:07}, due to conjugacy,
we can analytically marginalize the beta process measure $B$ and
obtain a predictive distribution simply over the concept assignments
$\mathbf{\hat{c}}^d$.  Taking the concentration parameter to be $b =
1$ yields the Indian buffet process (IBP) of Griffiths and Ghahramani~\cite{GriffithsGhahramani:05}.

The IBP is a culinary metaphor that describes how the sparsity
structure is shared across different draws of the Bernoulli process.
Each document is represented as a customer in an Indian buffet with
infinitely many dishes, where each dish represents a concept.  The
first customer (document) samples a $Poisson(\alpha_\omega)$ number of
dishes.  The $d$th customer selects a previously tasted dish
$j$ with probability $m_j/d$, where $m_j$ is the number of customers
to previously sample dish $j$.  He then chooses a
$Poisson(\alpha_\omega/d)$ number of new dishes.  With this metaphor,
it is easy to see that the sparsity pattern over concepts is shared
across documents because a document (customer) is more likely to pick
a concept (dish) if many previous documents have selected it.

\subsection{Nested Beta Process}
Above, we described how a beta process prior is well-suited to
modeling the presence or absence of concepts in each document $d$ of a
document collection.  However, we are still left with the task of
modeling the presence or absence of words in a particular concept.
Just as was the case at the concept level, a document should be more
likely to activate a word $i$ in a concept $j$ if many other documents
also have word $i$ active in concept $j$.

This leads to a natural extension of the IBP culinary metaphor to
include condiments that are added alongside each dish.  Specifically,
after the $d$th customer (document) selects her dishes (concepts) from the Indian buffet, she
selects an assortment of \emph{chutneys} (words) to accompany each
dish.  Analogous to what happens at the concept level, if a customer
is the first to sample from dish $j$, then she selects 
$Poisson(\alpha_\gamma)$ types of chutney.  The $d$th customer to
sample from dish $j$ selects chutney $i$ with probability $m_{ji}/d$,
where $m_{ji}$ is the number of previous customers who sampled chutney
$i$ alongside dish $j$.  She then selects $Poission(\alpha_\gamma/d)$
types of new chutney. 

\begin{figure*}[t]
\begin{center}
\subfigure{\includegraphics[width=6.5cm]{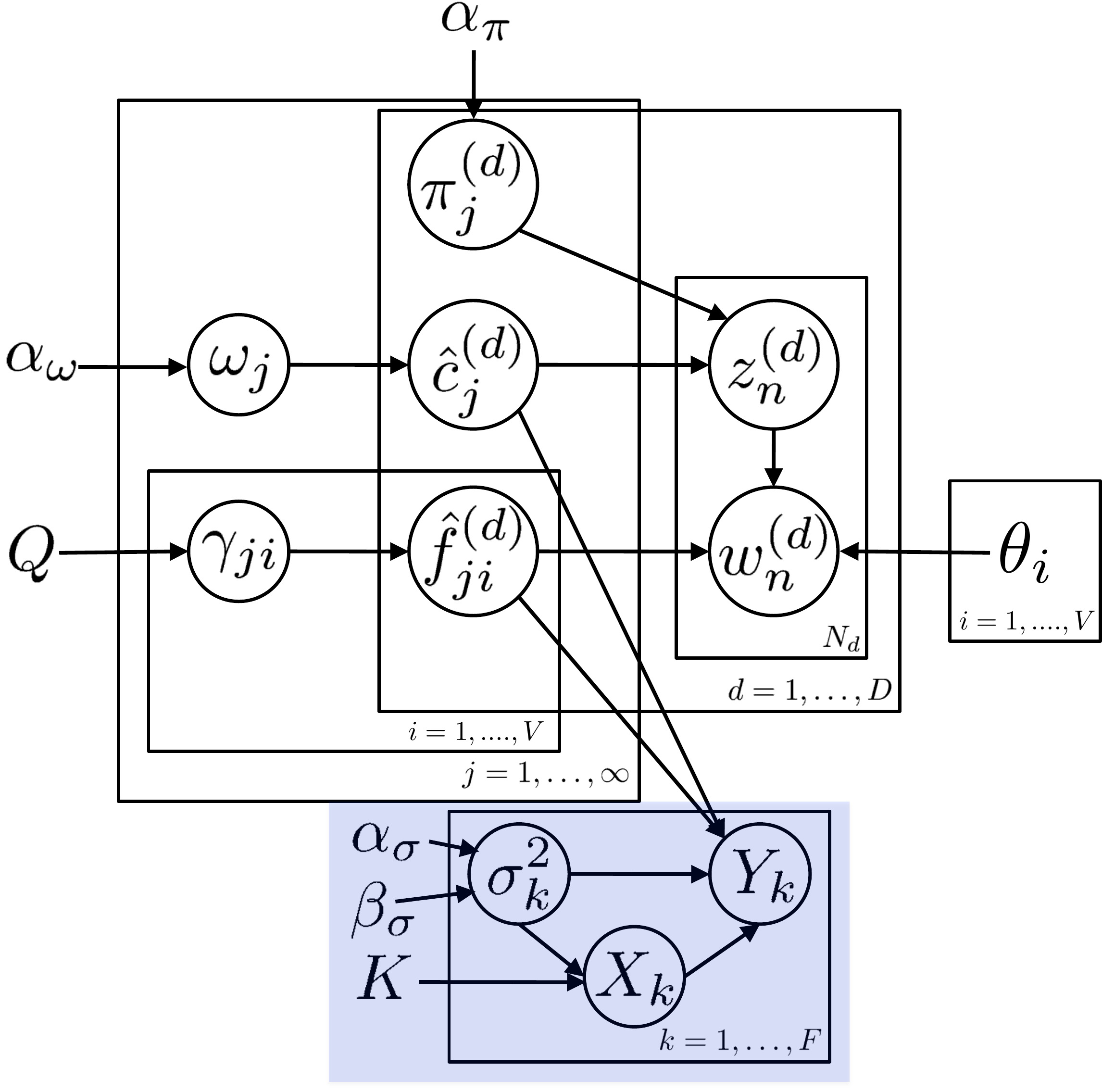}}
\subfigure{}{
\begin{minipage}[b]{0.5\linewidth}
\begin{algorithmic}
\footnotesize
\STATE \textbf{Given:} $Q = \sum_i \lambda_i \delta_{q_i}$, where $\lambda_i \in [0,1]$ and $q_i$ come from our vocabulary.
\STATE $(\boldsymbol{\omega}, \boldsymbol{\gamma}) \sim nBP(\alpha_\omega, Q)$
\FORALL{documents $d=1,\ldots,D$}
  \FORALL{concepts $j=1,\ldots$}
    \STATE $\piD{j} \sim Gamma(\alpha_\pi, 1)$
    \STATE $\cD{j} \sim Bernoulli(\omega_j)$
    \FORALL{$i=1,\ldots,V$}
      \STATE $\fD{ji} \sim Bernoulli(\gamma_{ji})$
    \ENDFOR
 \ENDFOR
 \FORALL{words $n$}
    \STATE $z_n^{(d)} \sim \boldsymbol{\pi}^{(d)} \odot \mathbf{\hat{c}}^{(d)} / \sum_{j:\cD{j} = 1} \piD{j}$
    \STATE $w_n^{(d)} | z_n^{\dpar} = z \sim \boldsymbol{\theta} \odot \mathbf{\hat{f}}^{(d)}_z / \sum_{i:\fD{z i} = 1} \theta_i$
  \ENDFOR
\ENDFOR
\FORALL{semantic features $k$}
  \STATE $\sigma_k^2 \sim InvGamma(\alpha_\sigma, \beta_\sigma)$
  \STATE $X_{kj} \sim N(0, \sigma_k^2/K)$, where $K$ is a positive scalar
  \STATE $Y_k^{T} \sim N((X_k \Phi)^T, \sigma_k^2 I_V)$
\ENDFOR
\end{algorithmic}
\end{minipage}
}
\vspace*{-5mm}
\caption{\footnotesize Plate diagram and generative model. We write
$(\boldsymbol{\omega}, \boldsymbol{\gamma}) \sim nBP(\alpha_\omega,
Q)$ for the distribution over the coin biases.\label{fig:model}}
\vspace*{-5mm}
\end{center}
\end{figure*}

Formally, this results in a \emph{nested} beta process (nBP), 
\begin{equation}
	B \sim \mbox{BP}\left(b_1,\mbox{BP}(b_2,B_0)\right),
\end{equation}
which can equivalently be described as ,
\begin{align}
	B = \sum_{j=1}^\infty \omega_j\delta_{B^*_j}, \quad B^*_j &= \sum_{i=1}^\infty \gamma_{ji}\delta_{\theta_{ji}}.\label{eqn:nbpPair}
\end{align}
That is, a draw from a nBP is a discrete measure whose atoms are
themselves discrete measures.  Just as Bernoulli process draws from
the top-level beta process results in concept assignments
$\hat{c}_j^{(d)}$, word assignments for concept $j$ are obtained by sampling
$\hat{f}_{ji}^{(d)} \sim Bernoulli(\gamma_{ji})$ from the lower level
beta process $B^*_j$, where $\hat{f}_{ji}^{(d)}$ indicates whether word $i$ is 
active in concept $j$ for document $d$ .  
A related idea of nesting beta processes is discussed in~\cite{Jordan:10}.

It is important to note that, in analogy to hierarchical clustering (e.g., the nested Dirichlet process of~\cite{Rodriguez:08}), the nBP is 
only weakly identifiable in that various combinations of dish-chutney
probabilities lead to the same likelihood of the data.  
Thus, only the prior specification differentiates these entities (e.g., encouraging fewer 
concepts each with more words or more concepts each with fewer words.)  Hierarchical clustering models often propose identifiability 
constraints such as the fact that components that correspond to the same class should be closer to each other than to components 
corresponding to other classes.  In our model of Sec.~\ref{sec:model}, we avoid such explicit constraints and instead 
incorporate global information which reduces the sensitivity of the model to nBP hyperparameter settings.  We do so in a way that 
maintains exchangeability (and thus computational tractability) of the model. 

\section{Capturing Superwords with Nested Beta Processes} 
\label{sec:model} 
Given a document collection of $D$ documents and a vocabulary of $V$
words, we wish to model the concepts therein as superwords.  In
particular, this entails identifying which concepts $j$ are present in
each document $d$ (indicated by $\cD{j} = 1$), along with which words
$i$ are active in concept $j$ for document $d$ (indicated by
$\fD{ji}=1$).  These binary variables are the nested beta process
features described in Sec.~\ref{sec:nestedBP}, with
$\mathbf{\hat{c}}$ representing the chosen dishes and
 $\mathbf{\hat{f}}$ the chosen chutneys. In particular,
we assume a prior $B \sim BP(1,BP(1,Q))$, with a discrete base measure
$Q = \sum_{i=1}^V \lambda_i \delta_{q_i}$, where $\lambda_i \in [0,1]$
and $q_i$ come from our vocabulary. 

Together, $\mathbf{\hat{c}}$ and $\mathbf{\hat{f}}$ define the makeup
of the concepts and their presence in each document.  However, to tie
these superwords to the actual document text, we must specify a
generative process for the observed words.  First, we model the
relative importance of each concept $j$ in a document $d$ as
i.i.d. gamma-distributed random variables, $\piD{j} \sim
Gamma(\alpha_\pi, 1)$.  We then associate the $n$th word in each
document with a concept assignment \znD, drawn from a multinomial
distribution proportional to $\boldsymbol{\pi}^{(d)} \odot
\mathbf{\hat{c}}^{(d)}$ (where $\odot$ refers to the element-wise
Hadamard product).\footnote{The non-zero elements of the resulting normalized distribution
are Dirichlet distributed, with dimensionality determined by the
number of ones in $\mathbf{\hat{c}}$.}  As such, \znD\ can only take values $j$ where
$\cD{j}=1$.  Likewise, given an assignment $\znD = j$, a document
generates its $n$th word, \wnD\, from a multinomial distribution
proportional to $\boldsymbol{\theta} \odot \mathbf{\hat{f}}^{(d)}_j$.
Here, $\theta_i$ is a parameter of our model indicating the relative
importance of words.\footnote{Alternatively, $\theta_i$ can be modeled
  similarly to $\piD{j}$, as gamma-distributed i.i.d. random
  variables.}  Our featural based model decouples concept presence in
a document from its prevalence.  Additionally, concepts that select
overlapping sets of words need not have the same marginal probability
of the shared word(s). A visual depiction of the graphical model as
well as a summary of the full generative process is provided in
Figure~\ref{fig:model}.

Our model as presented thus far suffers from the weak identifiability
issues described in Sec.~\ref{sec:nestedBP}.  In particular, the
representational flexibility of the model can lead to many concepts, each
containing few words, or a few concepts, each with many words, with
little difference in the likelihood of the data between the two cases.
We address this problem by incorporating semantic information about
our vocabulary in the manner described below.  (cf. the supplemental
material for an illustration of this problem on synthetic data.)

\paragraph{Incorporating semantic knowledge.}
We assume each word $i$ in our vocabulary is associated with an
\emph{observed}, real-valued feature vector in an $F$-dimensional semantic space.
Our fundamental assumption is that words appearing together in a 
concept should have similar semantic representations.  

Beyond addressing identifiability concerns, explicitly modeling
semantic features of our vocabulary allows us to elegantly incorporate
a variety of side information to help guide the makeup of concepts.
For instance, while the simplicity of the bag-of-words document representation 
provides many benefits in terms of computational efficiency, much
structure is thrown away that could be useful to particular retrieval
tasks. As such, if we want concepts to consist of words with
related or synonymous meanings, we might consider to use
sentence co-occurrence counts (e.g., in how many sentences do words $w_1$
and $w_2$ appear together?) as the basis for our semantic features.
There is a long history of work in linguistics studying such distributional 
similarity, popularized by R. F. Firth, who stated that ``you
shall know a word by the company it keeps,''~\cite{Firth:57}.
Rather than ignore this structural information, we can incorporate it
as part of our semantic feature set, as we do in the experiments we
consider in Sec.~\ref{sec:congress}.

We model this idea by assuming that \emph{concepts} are associated with
\emph{latent} semantic features in the same $F$-dimensional space.
Words that often co-occur in concept $j$ are expected to have semantic
representations that are well-explained by the features for concept
$j$. More formally, we consider an observed random matrix $Y$ of
dimensions $F \times V$, where each column $Y_{\bullet i}$ corresponds
to the semantic feature vector for word $i$.  Likewise, $X$ is a
random matrix with $F$ rows and a countably infinite number of
columns, one per concept.  As words can simultaneously be active in
multiple concepts (e.g., ``jaguar'' can refer to both a car and an animal), we assume that the expected value of the feature
representation for word $i$, $E[Y_{\bullet i}]$, is equal to a
weighted average over $X_{\bullet j}$ for all concepts $j$, with
weights proportional to the number of documents $d$ with $\fD{ji} =
1$.  In matrix notation, we write, $E[Y] = X \Phi$, where the
weight matrix $\Phi$ is deterministically computed from all
$\mathbf{\hat{c}}$ and $\mathbf{\hat{f}}$, such that $\Phi_{ji} =
\sum_{d:\cD{j} = 1} \fD{ji} / \sum_j \sum_{d:\cD{j} = 1} \fD{ji}$.
\footnote{It is important to note that in order to incorporate the semantic
information into our generative model, we rely on the fact that our
concepts are explicitly represented as \emph{sparse} sets of words.  The inclusion
or exclusion of a word in a concept directly informs which concepts
are responsible for the semantic meaning of that word.  To incorporate
the same information in traditional topic modeling approaches, we
would have all topics contributing to the semantic meaning of every word.}

By assuming independence of features and Gaussian noise, we can then specify the
generative distribution for $Y$ as described in
Figure~\ref{fig:model}.  In particular, we place conjugate normal and inverse gamma
priors on the latent concept features $X$ and the variance terms
$\sigma_k^2$, respectively, allowing us to analytically marginalize
them out for inference.

\paragraph{Example: Learning multilingual concepts from images.}
There are no modeling restrictions on the semantic data other than we
expect the features to be real-valued and with zero mean.  Hence, we
can take advantage of this flexibility to model semantics in a variety
of different forms, not limited to simply textual features.  To
demonstrate this flexibility, we consider the following toy problem:
given a small collection of dessert recipes in English and German,
downloaded from food blogs, can we learn concepts that are coherent
\emph{across} the two languages?  Based on image-based semantic
features, we address this problem without relying on parallel corpora
or an explicit dictionary as in the multilingual topic modeling
of~\cite{polylingual,multilingual}.  Specifically, assuming we have
an image associated with each word in our vocabulary (e.g., from a
Google images search), our semantic feature model encourages all
concepts $j$ that choose to include word $i$ to have latent feature
vectors that are similar to the image-based feature vector associated
with word $i$. We hypothesize that, despite a lack of co-occurrence between the
two languages within this small corpus, we should still obtain
reasonable multilingual concepts, because an apple looks like an
\emph{apfel}, eggs look like \emph{eier}, and so on.

Specifically, for each of 125 English and German vocabulary words, we
first collect the top three search results from Google Images.\footnote{Images for
    English words were retrieved from Google.com, and German words
    from Google.de, to avoid any internal translation that Google
    might otherwise do.}  We transform these images following the
  approach of Oliva and Torralba~\cite{Oliva:01}, to get simple,
  10-dimensional GIST-based features for each word, which we use as the
semantic features. We then run our sampling procedure from Sec.~\ref{sec:sampling} for
  10,000 samples, which we use to infer the marginal probabilities
  of any two words being active in the same concept.

\begin{figure}[t]
\begin{center}
\includegraphics[width=5cm]{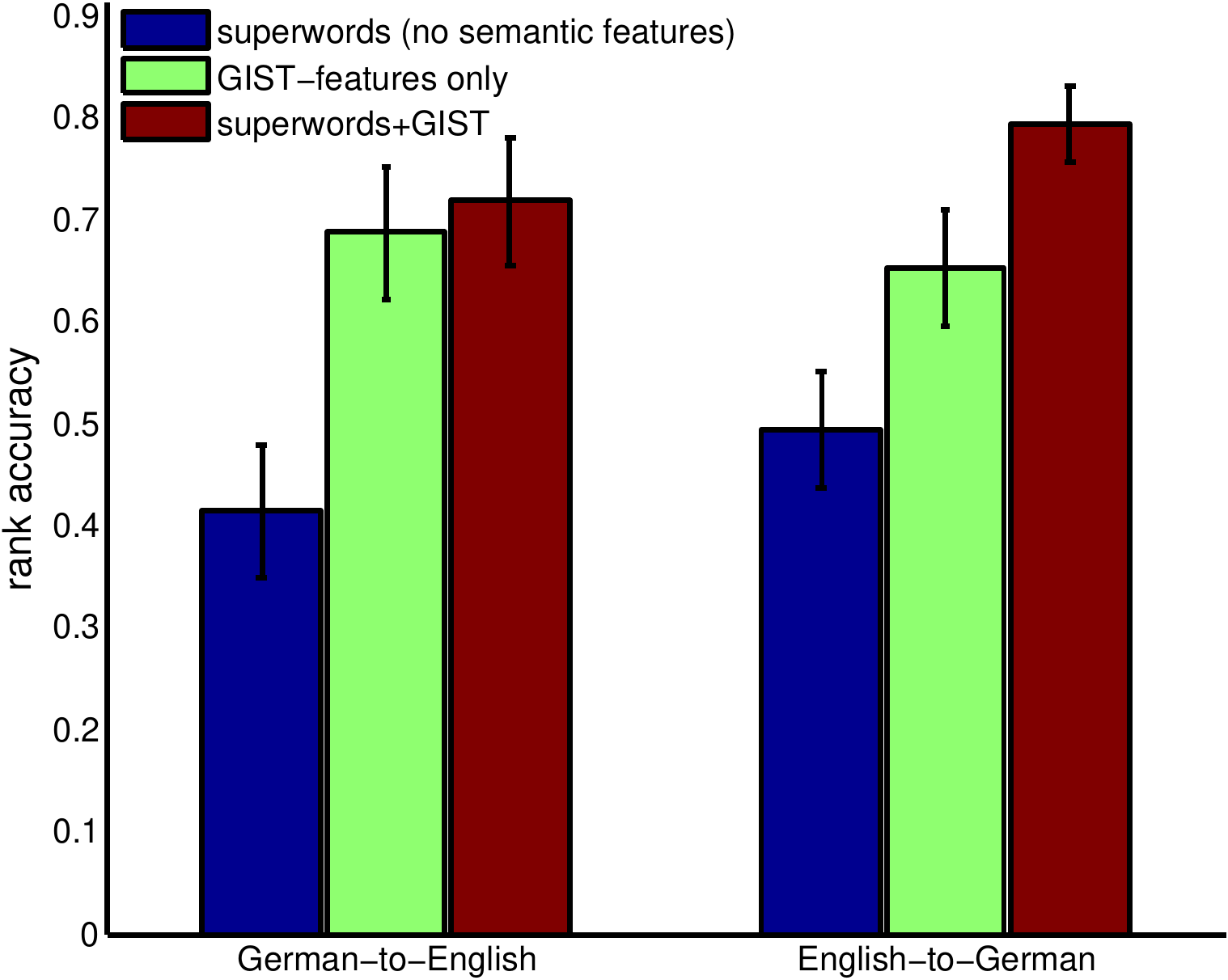}
\caption{\footnotesize{Rank accuracy of multilingual concepts inferred from
  English and German recipes; our model, combining image and corpus data,
outperforms predictions based on either text or images alone.\label{fig:rankAcc}}}
\end{center}
\end{figure}
 In order to quantitatively verify our hypothesis, we take this
marginal probability matrix and, for each word $w_i$, rank all other words
by probability of co-occurrence with $w_i$.  We then create a ground
truth set by finding all pairs of
words in our vocabulary that are considered English-German translations of each other according to
Google Translate.  We come up with 18 such synonym pairs.  We expand
each set to include direct synonyms that differ only in case or
number, giving us pairs of sets like $\{egg, eggs\} \leftrightarrow
\{eier\}$.  For each such pair, if $w_g$ is a German
translation of $w_e$, we see how high $w_e$ is ranked on $w_g$'s
marginal probability ranking, and compute the \emph{rank accuracy},
which is the percentage of the words $w_e$ is ranked higher than.  We
do this in both directions (from German to English and vice versa) for
all 18 set-pairs.  For comparison, we also compute rank accuracy when, rather
than using our model, we rank words directly based on their
$L_2$-distance in the 10-dimensional GIST feature space.
Figure~\ref{fig:rankAcc} shows that our model, combining corpus-based
concept modeling with simple image-based semantic features improves
performance over using the image features alone.  Of
course, as we expect, if we use our model without incorporating the images, we achieve poor performance, as
there is little cross-language information in the text.

Anecdotally, we find several concepts that do in fact represent
the types of cross-language coherence that we hoped for.  For example,
one concept consists primarily of kitchen tools and utensils, \{ whisk, rubber spatula,
baking sheet, butter, \emph{tortenheber} (cake server), \emph{schneebesen}
(whisk), \emph{weizenmehl} (wheat flour), \emph{mehl} (flour) \},
while another is heavy on mostly white and dry ingredients, \{
cornstarch, sugar, sugars,
\emph{zucker} (sugar), \emph{zuckers} (sugars), \emph{vanillezucker}
(vanilla sugar), \emph{salz} (salt), \emph{ricotta-kuchen} (ricotta
cake), ricotta, butter, \emph{p\"{u}rierstab}, \emph{teig} (dough),
\emph{mandel-zitronen-tarte}, \emph{ofen}, springform \}.  While
neither of these concepts is perfect, they are still indicative of the
flexibility and power of our semantic representation in guiding
concept formation.

\section{MCMC Computations} 
\label{sec:sampling}
In order to perform inference in our model, we employ a Markov chain
Monte Carlo (MCMC) method that interleaves Metropolis-Hastings (MH)
and Gibbs sampling updates.  Given that we are primarily interested in
the concept definitions ($\mathbf{\hat{f}}$) and their existence and
prevalence in each document ($\mathbf{\hat{c}}$ and
$\boldsymbol{\pi}$), we marginalize out all other random variables,
and end up with a collapsed sampler.  This marginalization is
analytically possible due to the conjugacy that exists throughout our
model.  Such collapsing is particularly critical to the
performance of our sampler given the number of binary indicator
variables in our model.  For instance, if we sample \cD{j}\ without
having marginalized out $\mathbf{z}$, we are forced to keep it set to
1 as long as any word $n$ in document $d$ has $z_n^{\dpar} = j$.

Another important consideration is that, while our model represents an infinite
number of concepts, only a finite number are ever instantiated at any given
time.  Thus, due to context-specific independence in our graphical model,
we find that the variables \fvv{j}{\dpar}\ and \piD{j}\ can be pruned from
the model for cases where $\cD{j} = 0$, and sampled only on an as-needed basis.

At a high level, our sampling approach is as follows:
\begin{algorithm}
\begin{algorithmic}[1]
\STATE Sample $\cvD |  \wvD, \pivD, \fvD, \cv{-d}, Y$ for every
  document $d$ using an MCMC procedure that proposes births/deaths for
  unique concepts.
\STATE Sample $\fD{ji} |  \wvD, \pivD, \cvD, \fnotD{(ji)}, Y$ for every
  document $d$, and every concept $j$ present in $d$, and every word $i$
  in the vocabulary.
\STATE Impute $z_1^{\dpar}, \ldots, z_{N_d}^{\dpar} | \wvD, \cvD, \fvD,
  \pivD$.
\STATE Use the imputed $\mathbf{z}$ variables to aid in sampling $\pivD 
  | \mathbf{z}^{\dpar}, \cD{j}, \pivv{-j}{\dpar}$.
\end{algorithmic}
\caption{\footnotesize High-level MCMC inference procedure\label{alg:sampler}}
\end{algorithm}

\vspace*{-5mm}
 For notational convenience, we write out two likelihood
terms for sampling \cvD\ and \fvD.  First,
\begin{align}
\lefteqn{P(\mathbf{w}^{\dpar} | \cvD, \boldsymbol{\pi}^{\dpar}, \fvD)
  =} \nonumber\\
& \left (\sum_{k: \cD{k} = 1} \piD{k}  \right )^{-N_d} \prod_{w \in \mathrm{doc} \; d} \left ( \sum_{z: \cD{z}=1 \wedge \fD{zw}=1} \frac{\theta_{w} \piD{z}}{\sum_{l: \fD{zl} = 1} \theta_{l}} \right )^{\wCountD{w}},
\end{align}
where \wCountD{w}\ counts the occurances of word $w$ in document $d$, and $N_d$
is the number of words in document $d$.  Second, by marginializing $X$
and $\sigma_k$ and recalling that $\Phi$ is determined from
$\mathbf{\hat{c}}$ and $\mathbf{\hat{f}}$,
\begin{equation}
P(Y|\Phi) \propto K^{|\matC|F/2} |\Phi_{\matC\cdot}\Phi_{\matC\cdot}^T +
  KI_{|\matC|}|^{-F/2} \prod_{k=1}^F \hat{\beta}_k^{-(\alpha_\sigma + V/2)},
\end{equation}
where $\matC$ is the set of active concepts, i.e., $\matC = \{ j : \sum_{d=1}^D \cD{j} > 0 \}$,  and, 
\begin{align}
\hat{\beta}_k &= \frac{1}{2}(Y_{k\cdot}Y_{k\cdot}^T - Y_{k\cdot}\Phi_{\matC\cdot}^T (\Phi_{\matC\cdot}\Phi_{\matC\cdot}^T +KI_{|\matC|})^{-1}\Phi_{\matC\cdot}Y_{k\cdot}^T) + \beta_\sigma.
\end{align}

The full derivations for the conditional distributions below can be found
in the supplementary material.

\paragraph{Sampling concept assignments to documents \cvD.}
In the case where we are sampling a concept that is shared among other
documents in the corpus (i.e., $\hat{c}_j^{(e)} = 1$ for some $e \neq
d$), we can sample from the conditional distribution:

\begin{align}
P(\cD{j} |  &\wvD, \pivD, \fvD, \cnotD{j}, \cvv{j}{(-d)}, Y)
  \propto \nonumber \\
& P(\cD{j} |
\cvv{j}{(-d)}) P(\mathbf{w}^{\dpar} | \cvD, \boldsymbol{\pi}^{\dpar}, 
\fvD) P(Y|\Phi). \label{eqn:cUpdate}
\end{align}

By IBP exchangeability, we assume that the current document is the last one, allowing us to write the 
prior probability on $\cD{j}$ (the first factor in this expression) as 
$P(\cD{j} | \cvv{j}{(-d)}) = m_j^{(-d)} / D$, where $m_j$ is the number of documents with $\cD{j} = 1$.

\begin{figure}[t]
\begin{center}
\subfigure[\label{fig:wpcHist}]{\includegraphics[width=2.3cm]{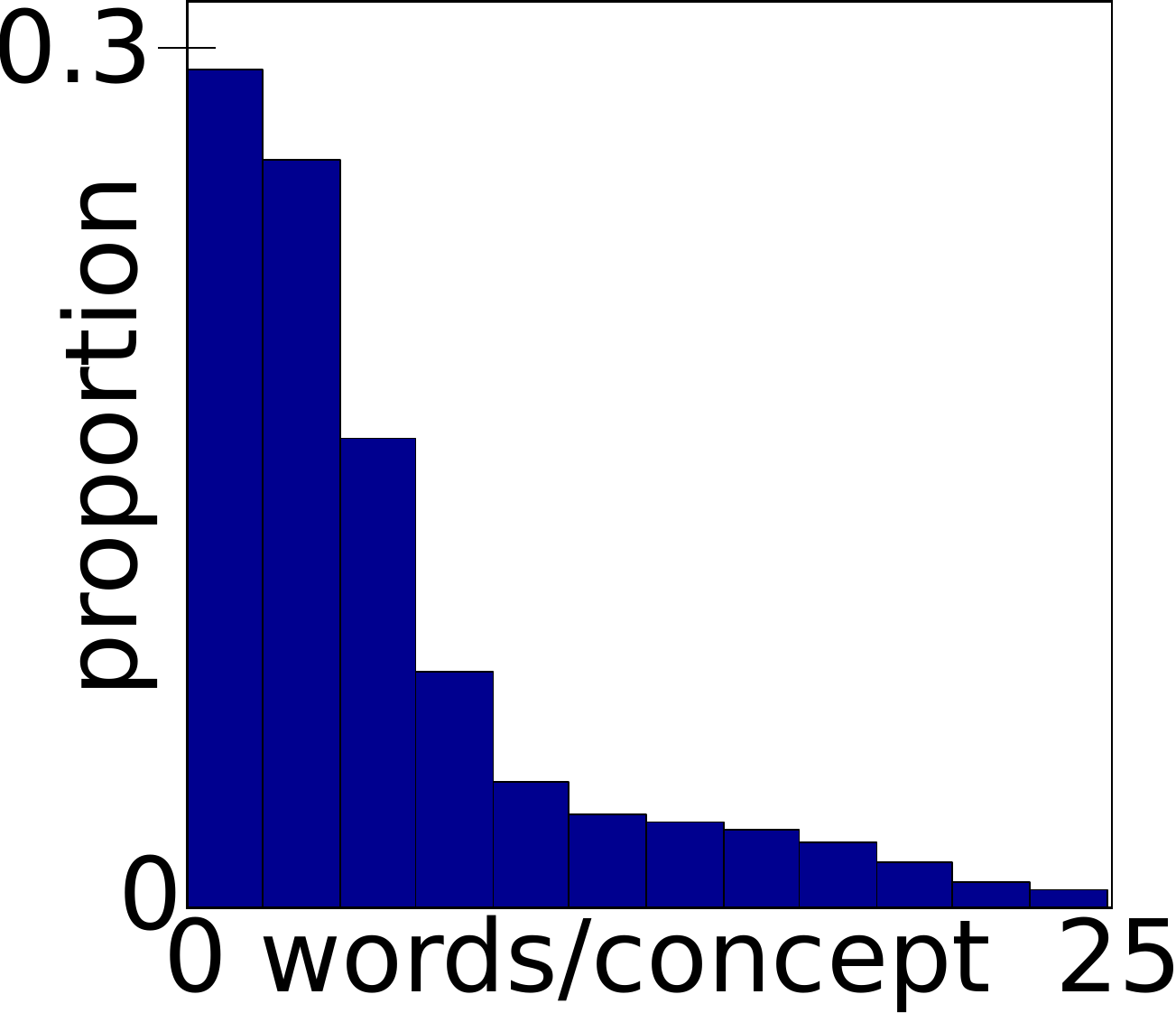}}
\subfigure[Democrat\label{fig:demWordle}]{\includegraphics[width=2.7cm]{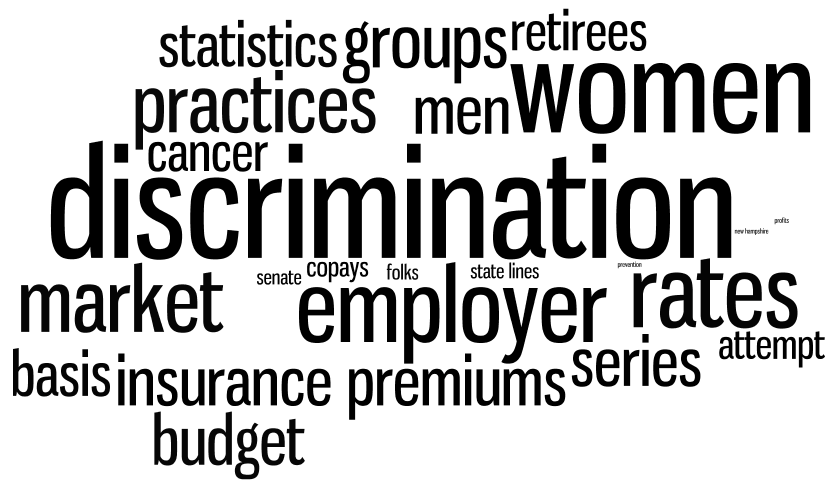}}
\subfigure[Republican\label{fig:repWordle}]{\includegraphics[width=2.7cm]{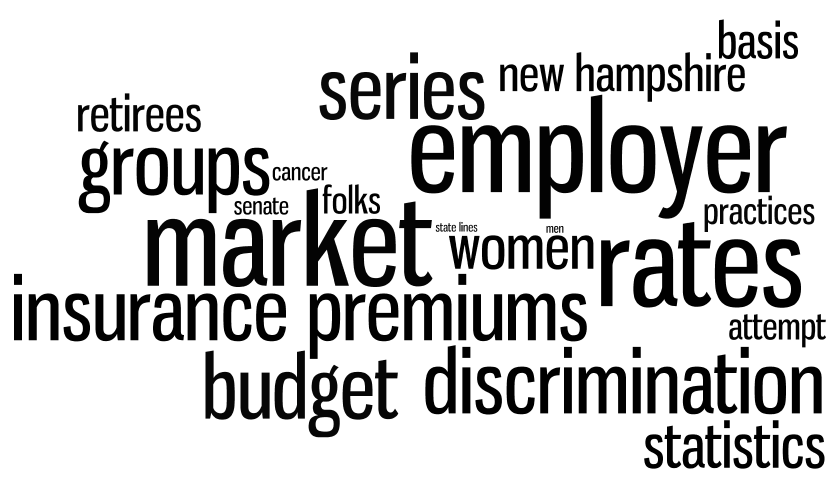}}
\vspace*{-5mm}
\caption{\footnotesize{\textbf{(a)} Histogram illustrating the
    word-level sparsity our model provides. \textbf{(b),(c)} Word
    clouds representing the same concept, but instantiated in
    Democratic and Republican members of Congress, respectively.\label{fig:wordles2}}}
\vspace*{-5mm}
\end{center}
\end{figure}

In order to sample unique concepts, we follow Fox et al.~\cite{Fox:NIPS09} and employ a birth/death
MH proposal.  More details can be found in the supplemental material.
\paragraph{Sampling concept definitions f}
This is similar to sampling a shared concept \cD{j}, except that the
domain (i.e., the vocabulary) is finite.  In particular,
we sample from the conditional 
\begin{align}
\lefteqn{P(\fD{ji} |  \wvD, \pivD, \cvD, \fnotD{(ji)}, Y) \propto}
\nonumber \\
& P(\fD{ji} | \fnotD{(ji)})
P(\mathbf{w}^{\dpar} | \cvD, \boldsymbol{\pi}^{\dpar}, \fvD) P(Y|\Phi) \label{eqn:fUpdate},
\end{align}

where, again by exchangeability, $P(\fD{ji} | \fnotD{(ji)}) =
(m_{ji}^{(-d)} + \lambda_i) / m_j$.  Here, $m_j$ is the number of
documents with $\cD{j} = 1$ and $m_{ji}^{(-d)}$ is the number of
documents $e \neq d$ assigning $\fvv{ji}{(e)} = 1$.

\paragraph{Imputing $\mathbf{z}^{\dpar}$ and Sampling
  $\boldsymbol{\pi}^{\dpar}$} This is straightforward and
details are left for the appendex.

\section{Empirical Results}
\label{sec:congress}

\begin{figure*}[t]
\begin{center}
\subfigure[\label{fig:w1}]{\includegraphics[width=4.5cm]{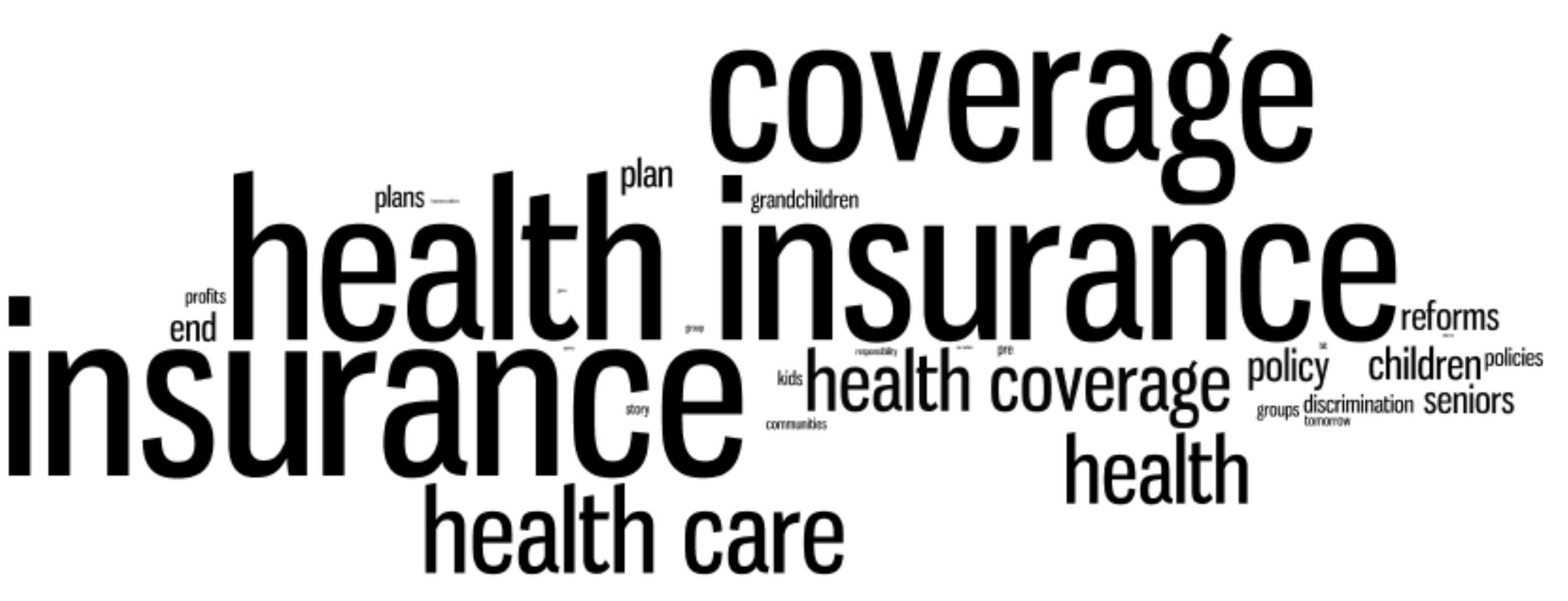}}
\subfigure[\label{fig:w2}]{\includegraphics[width=4.5cm]{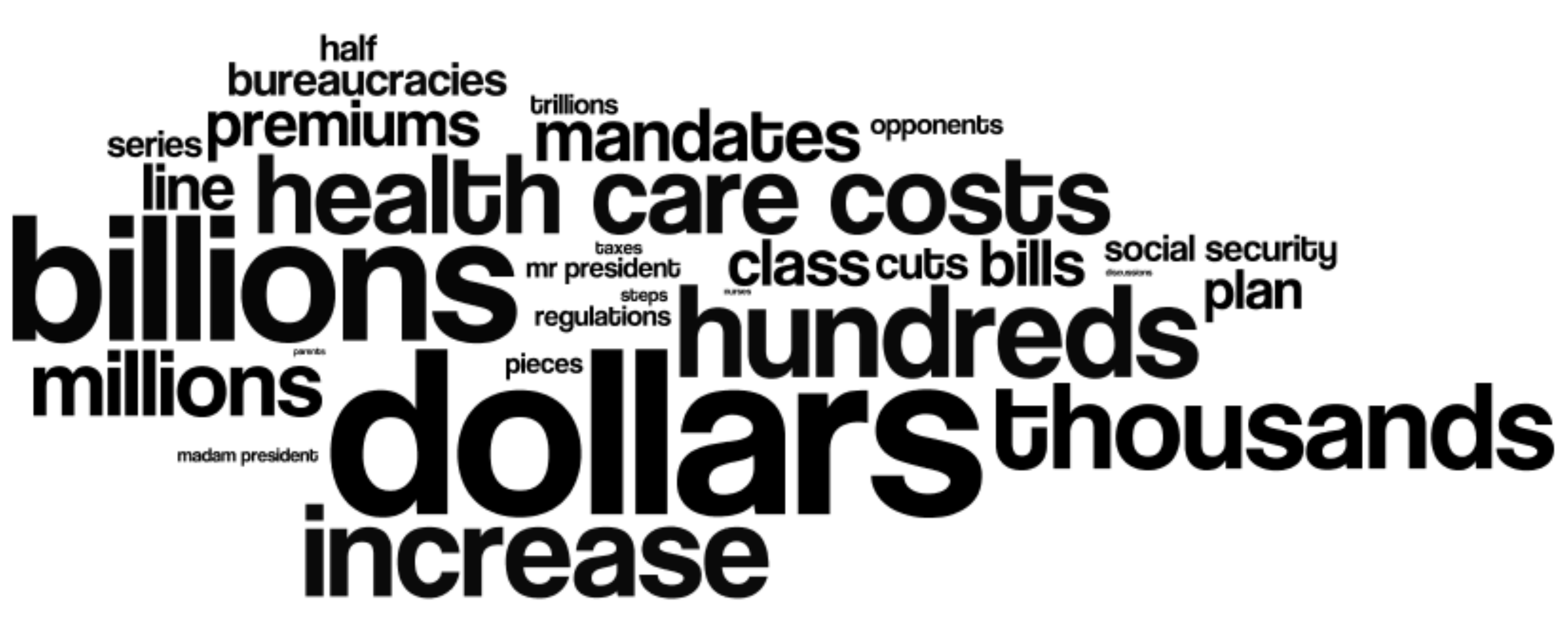}}
\subfigure[\label{fig:w3}]{\includegraphics[width=4.5cm]{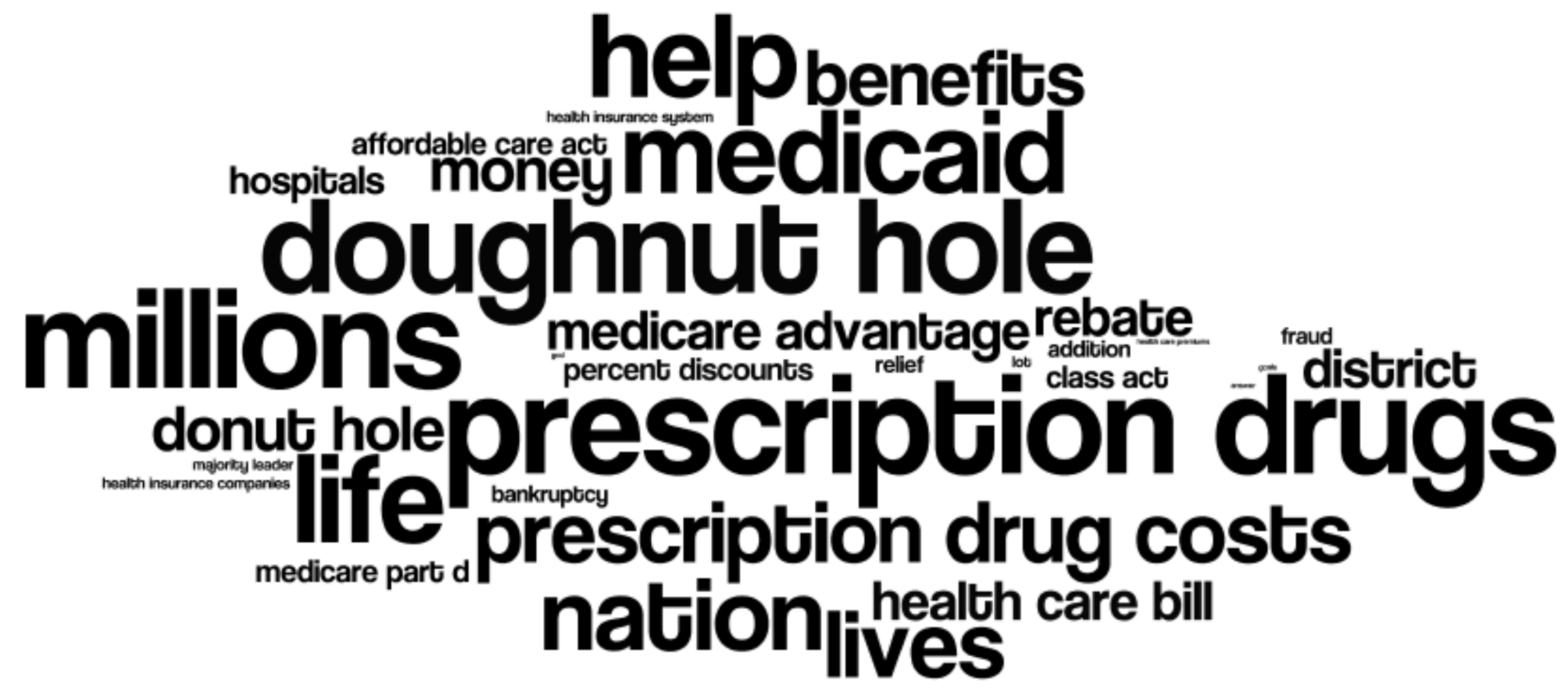}}
\vspace*{-5mm}
\caption{\footnotesize{Word clouds representing three concepts generated from our
  model, applied to the Congressional data set, using sentence
  co-occurrance-based semantic features.  \label{fig:wordles}}}
\vspace*{-5mm}
\end{center}
\end{figure*}

In order to evaluate how well our sparsity-inducing nested beta
process prior gives us the concept coherency and flexibility we desire, 
we compare concepts learned from our model to topics learned from an
HDP.  The HDP is a good model for comparison because it is widely used
as a probabilistic model for document collections with an unknown
number of topics, but does not model sparsity at either the topic or
word level.\footnote{The recent focused topic model (FTM) of
  Williamson et al.~\cite{FTM} provides concept-level sparsity, but
  topics are still distributions over the entire vocabulary.}
Our empirical evaluations in this section are all on a corpus
drawn from the Congressional Record during the health care debate that
marked the first two years of Barack Obama's presidency.  Each
document here represents a member of Congress (either in the Senate or
the House), and is a compilation of their statements from the chamber
floor during a few days of debate.  For analysis, we selected the
documents corresponding to the 100 most active speakers, as measured
by length of their floor transcripts.  After standard stop word
removal, we extracted named entities and noun phrases~\cite{stanfordTagger}, and removed
tokens that did not appear in at least 10 of the 100 documents,
leaving us with a vocabulary size of 682.  We augmented this corpus
with 50 semantic features, computed by taking a sentence-level word
co-occurrence matrix and projecting it to the top 50 principal
components (which account for approximately 60\% of the variance).

We run two parallel chains of an HDP sampler~\cite{ref:hdpCode} for
5,000 iterations, and choose the sample with the maximium joint
likelihood across both chains for our comparisons.  We also run two
parallel chains of our inference procedure from
Sec.~\ref{sec:sampling}, each to 1,000 samples, and again select the
most probable sample.\footnote{In the
supplemental material, we describe more details on how we initialize
our sampler, hyperparameter settings, and modifications to our
sampling procedure necessary for running efficiently on larger
corpora.}

As illustrated in Figure~\ref{fig:hdp-nbp-compare} and
previously discussed in the introduction, the concepts our model finds
are more coherent and focused than the topics learned by HDP.  The
word sparsity of our concepts is evident in the histogram of
Figure~\ref{fig:wpcHist}, showing that 99\% of the
empirical probability mass is on 25 concepts/word or fewer.
Figure~\ref{fig:wordles} shows three additional concepts that are
illustrative of the benefits of incorporating our sparsity-inducing
prior with semantic features.  For example, the concept found in
Figure~\ref{fig:w3} concisely represents the so-called donut hole (or ``doughnut hole,'' as we find
both spellings) in Medicare's prescription drug benefit, known as
Medicare Part D.

In order to measure this coherency in a more quantitative manner, we
conducted a user study where we presented participants with pairs of
word clouds corresponding to ten key ideas from the health care
debate (one cloud from our model and the other from HDP), and asked
them to choose the cloud from each pair that provides the more
coherent description of the word.  For fairness, we asked a public
policy expert to select ten words from our vocabulary representing the
most salient ideas in the debate, and used these words to generate the
word clouds.

For a given word, we found the HDP topic that assigned it the highest
probability, as well as the concept from our model that included it
the most times across the corpus.  The word clouds were generated
by removing the word in question and then displaying the remaining
words proportional to their weight.  To hide the identity of our
approach from the participants, we truncate each HDP topic to have the
same number of words as the corresponding concept from our model.
This gives the HDP topics an illusion of sparsity that they do not
naturally have, and hides one of the key advantages we have over topic
models.  For example, the ``Obamacare'' word clouds shown in
the top row of Figure~\ref{fig:hdp-nbp-compare} are transformed to the ones in the 
bottom row for the purpose of this study. Despite handicapping our model in this
manner, users found that our model produces more
coherent concept representations than the HDP.  Specifically, 73\% of
the 34 participants prefer our concepts to the HDP topics, with a mean
preference of 5.74/10.  Moreover, we note that the task itself
relied on a subtle understanding of American politics. For example,
overall, more participants preferred the HDP word cloud for
``Obamacare'' than ours, but when only considering participants who claimed
to have followed the health care debate closely (and presumably
understand that ``Obamacare'' is a term used pejoratively by Republicans),
this preference is flipped to ours.  More details on the study can be
found in the supplemental material.

Finally, since our model allows concepts to have different words
active in different documents, we can illustrate the flexibility of
our prior by creating different word clouds for the same concept, each
representing a subset of documents.  On this data set, a natural
separation of the documents is to have two partitions, one
representing Democrats and the other Republicans.
Figure~\ref{fig:wordles2} shows two word clouds from our
model representing the same concept, but one comes from Democrats and
the other from Republicans.  We can see the qualitative difference in
the word weights between the two populations, with ``discrimination''
and ``women'' being more active in this concept for Democrats while
``employer'' and ``market'' being more associated with Republicans.
We can find several similar anecdotes at the level of individual
documents.  For example, if we look at the leadership of the two
parties, we find that in one concept, the word ``Obamacare'' is
active for Republican Eric Cantor whereas ``health care bill''
is used for Democrat Chris Van Hollen.

\section{Discussion} 
\vspace{-0.05in}

Motivated by critical problems in information retrieval, in this
paper, we introduce a novel modeling technique for representing
concepts in document collections.  Popular generative models of text
have tended to focused on representing the ideas in a document collection
as probability distributions over the entire vocabulary, often leading
to diffuse, uninformative topic descriptions of documents.  In
contrast, our discrete, sparse representation provides a focused,
concise description that can, for example, enable IR systems to more
accurately describe a user's preferences.  A second key contribution
in this work is that our model naturally incorporates semantic
information that is not captured by alternative approaches; for
example, we demonstrated that images can be used to create
multilingual concepts, without any other translation information.   
Such side information can often be easily obtained, and help guide
topic models towards semantic meaningful predictions.  

Despite the promise, our model suffers from common limitations faced by many Bayesian nonparametric methods.  While
incorporating semantic features was helpful from a modeling
perspective, the matrix operations required to incorporate this data
can be expensive.  In experiments, we utilized 
parallelization and approximation techniques to reduce the running
time, but scalability remains a key aspect of future work. 

We believe that the notion of superwords,
characterized by sparse concepts and the incorporation of side
information through semantic features, can significantly improve the
effectiveness of IR techniques at capturing the nuances of users' preferences.   

\bibliographystyle{plainnat}  \small 
\bibliography{Bibliography}

\appendix
\section{Supplementary Material}
In this appendix, we provide detailed derivations of our MCMC sampling
procedure, as well as more details on the data we use for our experiments.

\subsection{MCMC Derivations}
In our sampling procedure, we assume that $\boldsymbol{\omega}$, 
$\boldsymbol{\gamma}$ and $\mathbf{z}$ are marginalized out of 
our model.  However, in order to sample $\mathbf{\hat{f}}$ 
and $\boldsymbol{\pi}$, we impute values for $\mathbf{z}$, which we discard at every
iteration.  As described in the main body of the paper, we employ a
Gibbs sampler that features birth/death Metropolis Hastings steps for
sampling $\mathbf{\hat{c}}$ and a Gibbs-within-Gibbs sampler for
jointly sampling $\mathbf{X}$ and $\mathbf{\hat{f}}$.

\subsubsection{Sample $\cvD | \wvD, \pivD, \mathbf{\hat{f}}, \cv{-d}, \mathbf{Y}$}
We assume that, at this moment, we have $J$ total concepts active across all documents.  $J^{(-d)}$ are 
active in all other documents, while $J_+^{(d)}$ are active only in this document 
(i.e., $J = J^{(-d)} + J_+^{(d)}$).  Without loss of generality, we assume that the concepts are numbered
such that the shared concepts ($1, 2, \ldots, J^{(-d)}$) come before the unique concepts 
($J^{(-d)}+1, \ldots, J$).

\paragraph{Sampling a shared concept}
Here, we consider sampling  $\cD{j} | \wvD, \pivD, \mathbf{\hat{f}},
\cnotD{j}, \cv{-d}, \mathbf{u}$ where $j \in \{1, \ldots, J^{(-d)} \}$,
i.e., concept $j$ is shared.
By Bayes' rule, as well as conditional independencies in the model, we have that:
\begin{align}
&P(\cD{j} | \wvD, \pivD, \mathbf{\hat{f}}, \cnotD{j}, \cv{-d}, \mathbf{Y}) \\ \nonumber
&\hspace{0.05in}\propto
P( \wvD, \mathbf{Y} | \mathbf{\hat{c}}, \pivD, \mathbf{\hat{f}}) \cdot P(\cD{j} | \cvv{j}{(-d)}) \\
&\hspace{0.05in}= P( \wvD | \cvD, \pivD, \fvD) \cdot P(\mathbf{Y} | \Phi) \cdot P(\cD{j} | \cvv{j}{(-d)})
\end{align}
By IBP exchangeability, we assume that the current document is the last one, allowing us to write the 
prior probability on $\cD{j}$ (the last factor in this expression) as 
$P(\cD{j} | \cvv{j}{(-d)}) = m_j^{(-d)} / D$, where $m_j$ is the number of documents with $\cD{j} = 1$.

The first factor, which we call the text likelihood term, can be simplified as follows:

\begin{align}
\lefteqn{P(\mathbf{w}^{\dpar} | \cvD, \boldsymbol{\pi}^{\dpar}, \fvD )} \\
&= \sum_{\mathbf{z}^{\dpar}} \prod_{n=1}^{N_d} P(w_n^{\dpar} | z_n^{\dpar}, \fvD) \cdot P(z_n^{\dpar} | \cvD, \pivD) \\
&=  \prod_{w \in \mathrm{doc} \; d} \left ( \sum_z \left ( \frac{\theta_{w} \fD{zw}}{\sum_l \theta_{l} \fD{zl}} \right ) \left ( \frac{\piD{z} \cD{z}}{\sum_k \piD{k} \cD{k}} \right )  \right )^{\wCountD{w}} \\
&\propto \left (\sum_{k: \cD{k} = 1} \piD{k}  \right ) ^ {-N_d}\hspace{-0.15in} \prod_{w \in \mathrm{doc} \; d} \left ( \sum_{z: \cD{z}\cdot\fD{zw}=1} \frac{\piD{z}}{\sum_{l: \fD{zl} = 1} \theta_{l} } \right )^{\wCountD{w}} 
\end{align}
where \wCountD{w}\ is the count of word $w$ in document $d$, and $N_d$ is the total word count for document $d$.  Note that if there is a word $i$ that appears in document $d$ such that the only concept that explains it is $j$ 
(i.e., $\fD{ji} = 1$ but $\fD{ki} = 0$ for all $k \neq j$), then \cD{j}\ must be set to 1.

The second factor, $P(\mathbf{Y} | \Phi)$, which brings in influence from the semantic
features, is derived at the end of this supplementary material.

We sample from this conditional distribution using a Metropolis Hastings step, where we have a 
deterministic proposal that flips the current value of \cD{j}\ from $c$ to $\bar{c}$.  Specifically, we
flip \cD{j}\ with the following acceptance probability:
\begin{align}
\rho(\bar{c}|c) = \min \left \{ \frac{P(\cD{j} = \bar{c} | \wvD, \pivD, \mathbf{\hat{f}},
\cnotD{j}, \cv{-d}, \mathbf{u})}{P(\cD{j} = c |\wvD, \pivD, \mathbf{\hat{f}},
\cnotD{j}, \cv{-d}, \mathbf{u})}, 1 \right \}.
\end{align}
Note that if we consider flipping \cD{j}\ from 0 to 1, we will need to first sample values 
for \fD{j}\ and \piD{j}\ from their priors, since they wouldn't otherwise exist.  We sample
\piD{j}\ from $Gamma(\alpha_\pi, 1)$ and sample \fD{j}\ from its prior in Equation~\ref{eqn:fPrior}.

\paragraph{Sampling unique concepts}
Let $\cvv{+}{\dpar}$ be the current unique concepts for document d,
and $\fvv{+}{\dpar}$ and $\pivv{+}{\dpar}$  be their associated
parameters.  (Let $\cvv{-}{\dpar}$, $\fvv{-}{\dpar}$ and
$\pivv{-}{\dpar}$ be the same for shared concepts.)

To sample these unique concepts, we'll use a birth/death proposal distribution, which factors as follows:
\begin{align}
q(\cvv{+}{\dpar'}, \fvv{+}{\dpar'}, \pivv{+}{\dpar'} |
\cvv{+}{\dpar}, \fvv{+}{\dpar}, \pivv{+}{\dpar})
&= q_c(\cvv{+}{\dpar'} | \cvv{+}{\dpar})\\ \nonumber
&\hspace{-1.5in} \cdot q_f(\fvv{+}{\dpar'} | \fvv{+}{\dpar}, \cvv{+}{\dpar'}, \cvv{+}{\dpar}) 
q_\pi(\pivv{+}{\dpar'} | \pivv{+}{\dpar}, \cvv{+}{\dpar'}, \cvv{+}{\dpar}).
\end{align}
In our IBP, the $D$th customer is supposed to sample $Poisson(\alpha_\omega / D)$ new dishes, and
therefore the probability of a concept birth (which we call $\eta(J_+^{\dpar})$) 
is the probability that such a draw would result in more than the current number of unique concepts, $J_+^{\dpar}$.  
In particular, we define $\eta(J_+^{\dpar})= 1 - PoissonCDF(J_+^{\dpar} ; \alpha_\omega / D)$, for $J_+^{\dpar} > 0$.  If there are no unique
concepts, then we are forced to propose a birth, and as such, $\eta(0) = 1$.

Thus, the proposal $q_c(\cvv{+}{\dpar'} | \cvv{+}{\dpar})$ adds a new concept with probability $\eta(J_+^{\dpar})$, and 
kills off each of the current $J_+^{\dpar}$ concepts with probability $\frac{1 - \eta(J_+^{\dpar})}{J_+^{\dpar}}$.
The proposals for $f$ and $\pi$ draw from their priors
($Bernoulli(\boldsymbol{\lambda})$ and 
$Gamma(\alpha_\pi, 1)$, respectively) for a concept birth, and otherwise
deterministically maintain the current values for the existing concepts.

Given this definition for our birth/death proposal, we can now write down our Metropolis Hastings step
for sampling the unique \cvv{+}{\dpar}\ and associated parameters.  In particular, we accept the 
Metropolis Hastings acceptance ratio is given by
%
%
\begin{align}
\lefteqn{r \left ( \cvv{+}{\dpar'}, \fvv{+}{\dpar'}, \pivv{+}{\dpar'} |
\cvv{+}{\dpar}, \fvv{+}{\dpar}, \pivv{+}{\dpar} \right ) =} \nonumber \\
&\frac{P \left ( \cvv{+}{\dpar'}, \fvv{+}{\dpar'}, \pivv{+}{\dpar'} 
| \wvD, \cvv{-}{\dpar}, \cv{-d}, \fvv{-}{\dpar}, \pivv{-}{\dpar},
\fvnotD, \mathbf{Y} \right )}
{P \left ( \cvv{+}{\dpar}, \fvv{+}{\dpar}, \pivv{+}{\dpar} | 
\wvD, \cvv{-}{\dpar}, \cv{-d}, \fvv{-}{\dpar}, \pivv{-}{\dpar},  \fvnotD, \mathbf{Y} \right )} \nonumber\\
& \cdot 
\frac{q \left ( \cvv{+}{\dpar}, \fvv{+}{\dpar}, \pivv{+}{\dpar} | 
\cvv{+}{\dpar'}, \fvv{+}{\dpar'}, \pivv{+}{\dpar'} \right )}
{q \left ( \cvv{+}{\dpar'}, \fvv{+}{\dpar'}, \pivv{+}{\dpar'} | 
\cvv{+}{\dpar}, \fvv{+}{\dpar}, \pivv{+}{\dpar} \right )}.
\end{align}

\noindent
Using Bayes' Rule and conditional independencies of our model, we can rewrite the first fraction as follows:
\begin{align}
&\hspace{-0.05in}\frac{P \left ( \wvD | [\cvv{-}{\dpar} \cvv{+}{\dpar'}],
    [\pivv{-}{\dpar} \pivv{+}{\dpar'}], [\fvv{-}{\dpar}
    \fvv{+}{\dpar'}] \right )}{P \left ( \wvD | [\cvv{-}{\dpar} \cvv{+}{\dpar}], [\pivv{-}{\dpar}
	    \pivv{+}{\dpar}], [\fvv{-}{\dpar} \fvv{+}{\dpar}] \right )} \cdot \nonumber\\
	&\frac{
P \left ( \mathbf{Y} | [\cvv{-}{\dpar} \cvv{+}{\dpar'}],  [\fvv{-}{\dpar}
    \fvv{+}{\dpar'}] \right ) P \left ( \cvv{+}{\dpar'} \right ) P
	  \left ( \fvv{+}{\dpar'} \right ) P \left ( \pivv{+}{\dpar'} \right )}{P \left ( \mathbf{Y} | [\cvv{-}{\dpar} \cvv{+}{\dpar}],  [\fvv{-}{\dpar}
	    \fvv{+}{\dpar}] \right )P \left ( \cvv{+}{\dpar} \right ) P \left ( \fvv{+}{\dpar} \right ) P \left ( \pivv{+}{\dpar} \right )} 
\end{align}
\noindent
Likewise, by definition, we can factor the second fraction as follows:
\begin{align}
\frac{q_c(\cvv{+}{\dpar} | \cvv{+}{\dpar'}) q_f(\fvv{+}{\dpar} | \fvv{+}{\dpar'}, \cvv{+}{\dpar'}, \cvv{+}{\dpar}) 
q_\pi(\pivv{+}{\dpar} | \pivv{+}{\dpar'}, \cvv{+}{\dpar'}, \cvv{+}{\dpar}) }{q_c(\cvv{+}{\dpar'} | \cvv{+}{\dpar}) q_f(\fvv{+}{\dpar'} | \fvv{+}{\dpar}, \cvv{+}{\dpar'}, \cvv{+}{\dpar}) 
q_\pi(\pivv{+}{\dpar'} | \pivv{+}{\dpar}, \cvv{+}{\dpar'}, \cvv{+}{\dpar}) }.
\end{align}
\noindent
Plugging in the corresponding terms, if we propose a birth, we can simplify the acceptance ratio to:

\begin{align}
&\hspace{-0.2in}r \left ( \cvv{+}{\dpar'}, \fvv{+}{\dpar'}, \pivv{+}{\dpar'} |
\cvv{+}{\dpar}, \fvv{+}{\dpar}, \pivv{+}{\dpar} \right ) = \notag \\
& \frac{P \left ( \wvD | [\cvv{-}{\dpar} \cvv{+}{\dpar'}],
    [\pivv{-}{\dpar} \pivv{+}{\dpar'}], [\fvv{-}{\dpar}
    \fvv{+}{\dpar'}] \right ) }{P \left ( \wvD | [\cvv{-}{\dpar}
    \cvv{+}{\dpar}], [\pivv{-}{\dpar} \pivv{+}{\dpar}],
    [\fvv{-}{\dpar} \fvv{+}{\dpar}] \right ) } \nonumber\\
&\cdot\frac{	Poisson(J_+^{\dpar} + 1; \alpha_\omega /
D) (1 - \eta(J_+^{\dpar} + 1))}{ Poisson(J_+^{\dpar};
\alpha_\omega / D) (J_+^{\dpar} + 1) \eta(J_+^{\dpar})}
\notag \\
&\cdot\frac{P \left ( \mathbf{Y} | [\cvv{-}{\dpar} \cvv{+}{\dpar'}],  [\fvv{-}{\dpar}
    \fvv{+}{\dpar'}] \right ) }{P \left ( \mathbf{Y} | [\cvv{-}{\dpar} \cvv{+}{\dpar}],  [\fvv{-}{\dpar}
    \fvv{+}{\dpar}] \right ) } \label{eqn:birthconcept}.
\end{align}

\noindent
Likewise, if we propose to kill a concept, we have exactly the same form for the acceptance ratio, replacing the third line by:
\begin{align}
\frac{Poisson(J_+^{\dpar} - 1; \alpha_\omega /
  D) J_+^{\dpar} \eta(J_+^{\dpar} - 1)}{
  Poisson(J_+^{\dpar}; \alpha_\omega / D) (1 - \eta(J_+^{\dpar})) }
\label{eqn:killconcept}.
\end{align}
\subsubsection{Impute $z_1^{\dpar}, \ldots, z_{N_d}^{\dpar} | \wvD, \cvD,
  \fvD, \pivD$}
\noindent
For $n = 1, 2, \ldots, N_d$:

\noindent
By Bayes' rule:
\begin{multline}
P(\znD = z | \wnD = w, \cvD, \fvD, \pivD) \propto \\
 P(\wnD = w | \znD = z, \fvD) \cdot P(\znD = z | \pivD, \cvD).
\end{multline}
\noindent
Note that if $\cD{z} = 0$ or $\fD{zw} = 0$, the probability of assigning $\znD = z$ is 0.
Therefore, when imputing the value of \znD (associated with word $w$), we only need to consider values $z$ such that
$\cD{z} = 1$ and $\fD{zw} = 1$.  In this case, we sample $z$ from:
\begin{align}
\left ( \frac{\theta_{w}}{\sum_{i:\fD{zi}=1} \theta_{i}} \right ) \left ( \frac{\piD{z}}{\sum_{j:\cD{j}=1} \piD{j}} \right ).
\end{align}
\noindent
Note that the numerator of the first factor and the denominator of the second factor are the same across all values of $z$, so we sample $\znD=z$ 
(for $z$ such that $\cD{z} = 1$ and $\fD{zw} = 1$) proportional to:
\begin{align}
\frac{\piD{z}}{\sum_{i:\fD{zi}=1} \theta_{i}}.
\end{align}
\subsubsection{Sample $\piD{j} | \zvD, \pinotD{j},  \cvD$}
\noindent
One can think of the concept frequency random variables $\piD{j}$ as utility random variables aimed at modeling the concept frequency \emph{distribution}
\begin{align}
\tilde{\pi}^{(d)} = \frac{\boldsymbol{\pi}^{(d)} \odot
	\mathbf{\hat{c}}^{(d)}}{\sum_j \piD{j}\cD{j}}.	
\end{align}
Specifically, only considering the non-zero components of $\tilde{\pi}^{(d)}$ (specified by the $\cD{j}=1$), this distribution is Dirichlet distributed and the gamma random variables are used in constructing a draw from this distribution.  The purpose of utilizing the $\piD{j}$ in place of working with $\tilde{\pi}^{(d)}$ directly is because the dimensionality of the underlying Dirichlet distribution is changing as $\hat{c}^{(d)}$ changes and thus we can maintain an infinite collection of gamma random variables that are simply accessed during the sampling procedure.

We have multinomial observations $\zvD$ (representing the word-concept assignments) from the concept frequency distribution $\tilde{\pi}^{(d)}$.  Due to the inherent conjugacy of multinomial observations to a Dirichlet prior, the posterior of $\tilde{\pi}^{(d)}$ is (using a slight abuse of notation):
\begin{align}
	\tilde{\pi}^{(d)} \mid \zvD, \mathbf{\hat{c}}^{(d)} \sim \mbox{Dir}([\alpha_\pi + n_1^{(d)}, \alpha_\pi + n_2^{(d)}, \dots] \odot \mathbf{\hat{c}}^{(d)}).
\end{align}

Once again, we can work in terms of our utility random variables $\piD{j}$ to form a draw from the desired Dirichlet posterior (again, along the non-zero components). Specifically, we draw
\begin{align}
\piD{j} | \zvD \sim Gamma(\piD{j}; \alpha_\pi + n_j^{(d)}, 1).
\end{align}
If $\cD{j} = 0$, necessarily we will not have any counts of concept $j$ in document $d$ (i.e., $n_j^{(d)} = 0$), implying that the distribution of these utility random variables remains the same.  Thus, in practice we only need to resample the utility random variables $\piD{j}$ for which $\cD{j} = 1$.
%
%

\subsubsection{Sample $\fD{ji} | \wvD, \fnotD{(ji)}, \fvnotD, \zvD,
  \mathbf{\hat{c}} $}
First, we note that \fvv{j}{\dpar}\ only exists in documents $d$ where $\cD{j} = 1$.  (If $\cD{j} = 0$,
it means that all observations \wvD\ are independent of \fvv{j}{\dpar}, and thus such $f$ nodes
can be pruned.)  Second, while, due to conjugacy, we can write the
conditional distribution for $\fD{ji}$ without using~\zvD\ (as was the
case when sampling $\mathbf{\hat{c}}$), we use the imputed ~\zvD\ 
here for computational efficiency.

We have the following:
\begin{align}
\lefteqn{P(\fD{ji} | \wvD, \fnotD{(ji)}, \fvnotD, \zvD,
  \mathbf{\hat{c}})}\\
 = &\int_0^1 P(\fD{ji}, \gamma_{ji} | \wvD, \fnotD{(ji)}, \fvnotD,
 \zvD, \mathbf{\hat{c}}) d\gamma_{ji} \\
= & P(\wvD | \fvD, \zvD) P(\mathbf{Y} | \mathbf{\hat{c}}, \mathbf{\hat{f}}) \int_0^1 P(\fD{ji}, \gamma_{ji} | \fvnotDji) d\gamma)_{ji}\\
\propto &\left ( \prod_{n=1}^{N_d} P(\wnD | \fvD, \znD) \right )  P(\mathbf{Y} | \mathbf{\hat{c}}, \mathbf{\hat{f}})  \int_0^1 P(\mathbf{\hat{f}}_{ji}^{(d:\cD{j}=1)}, \gamma_{ji}) d\gamma_{ji}.
\end{align}

The first factor in this expression is the corpus likelihood term, assuming $\znD$.  We can simplify this as follows:
\begin{align}
 \lefteqn{\prod_{n=1}^{N_d} P(\wnD | \fvD, \znD)}\\
&= \prod_{n=1}^{N_d} \frac{\theta_{ \wnD} \fD{\znD \wnD}}{\sum_{l}\theta_{l} \fD{\znD l}} \\
&\propto \prod_{n: \znD=j} \frac{\theta_{\wnD} \fD{j \wnD}}{\sum_{l}\theta_{l} \fD{jl}} \\
&= \left ( \sum_{l:\fD{jl} = 1}\theta_{l} \right )^{-\nD{j}} \prod_{n: \znD=j} \theta_{ \wnD} \fD{j \wnD}.
\end{align}

Recall that the second factor is the semantic likelihood term, and is derived at the end of this supplementary material.

Finally, the integral, representing the prior probability on $\hat{f}_{ji}$, is simplified as follows:
\begin{align}
\lefteqn{\int_0^1 P(\mathbf{\hat{f}}_{ji}^{(d:\cD{j}=1)}, \gamma_{ji}) d\gamma_{ji} }\\
&= \int_0^1 P(\mathbf{\hat{f}}_{ji}^{(d:\cD{j}=1)}| \gamma_{ji}) P(\gamma_{ji}) d\gamma_{ji} \\
&= \int_0^1 \prod_{d:\cD{j}=1} P(\fD{ji}| \gamma_{ji}) P(\gamma_{ji}) d\gamma_{ji} \\
&= \int_0^1 \gamma_{ji}^{m_{ji}} (1 - \gamma_{ji})^{m_j - m_{ji}} Beta(\gamma_{ji};  \lambda_i, 1 - \lambda_i) d\gamma_{ji} \\
&= \frac{1}{B( \lambda_i,   1 - \lambda_i)} \int_0^1 \gamma_{ji}^{m_{ji} +   \lambda_i - 1} (1 - \gamma_{ji})^{m_j - m_{ji} - \lambda_i} d\gamma_{ji} \\
&= \frac{B(m_{ji} +   \lambda_i, m_j - m_{ji} +   1 - \lambda_i)}{B(  \lambda_i,   1 - \lambda_i)} \\
&\propto \Gamma(m_{ji} +   \lambda_i) \Gamma(m_j - m_{ji} +  1 - \lambda_i),
\end{align}
where $m_{ji}$ is the number of documents with both $\cD{j} = 1$ and $\fD{ji} = 1$.  Thus, if $\fD{ji} = 1$, we have:
\begin{align}
\lefteqn{\Gamma(m_{ji}^{(-d)} +   \lambda_i + 1) \Gamma(m_j - m_{ji}^{(-d)}  - \lambda_i ) =}\\
&= (m_{ji}^{(-d)} +   \lambda_i) \Gamma(m_{ji}^{(-d)} +   \lambda_i)\Gamma(m_j - m_{ji}^{(-d)} +  \lambda_i),
\end{align}
and if $\fD{ji} = 0$, we have:
\begin{align}
\lefteqn{\Gamma(m_{ji}^{(-d)} +   \lambda_i) \Gamma(m_j - m_{ji}^{(-d)} +  1 - \lambda_i) =}\\
&= \Gamma(m_{ji}^{(-d)} +   \lambda_i)(m_j - m_{ji}^{(-d)}  - \lambda_i)\Gamma(m_j - m_{ji}^{(-d)} -\lambda_i).
\end{align}
After some cancellation, we get the following:
\begin{align}
P(\fD{ji} = 1 | \fvnotDji) &= \frac{m_{ji}^{(-d)} +   \lambda_i}{m_j}, \label{eqn:fPrior}\\
P(\fD{ji} = 0 | \fvnotDji) &= \frac{m_j - m_{ji}^{(-d)} - \lambda_i}{m_j }.
\end{align}

To recap, we sample \fD{ji}\ as follows:
\begin{enumerate}
\item if $\nD{j} = 0$, we sample $\fD{ji}$ proportional to $ P(\mathbf{Y} | \mathbf{\hat{c}}, \mathbf{\hat{f}})  P(\fD{ji} | \fvnotDji)$.
\item if $\nD{j} > 0$ but $\nD{ji} = 0$, we sample proportional to
 $ P(\mathbf{Y} | \mathbf{\hat{c}}, \mathbf{\hat{f}}) P(\fD{ji} | \fvnotDji) \left ( \sum_{l:\fD{jl}=1}\theta_{l} \right )^{-\nD{j}}$.
\item else, if $\nD{j} > 0$ and $\nD{ji} > 0$, set $\fD{ji} = 1$ with probability 1.
\end{enumerate}

\subsubsection{Determine $p(\mathbf{Y} | \Phi)$}

Recall that $F$ denotes the dimensionality of our semantic features.  If the number of concepts were finite with $J$ concepts, we could specify
\begin{align}
       X &\mid \Sigma \sim MN(M,\Sigma,K)\\
       \Sigma &\sim \mbox{IW}(n_0,S_0),
\end{align}
where $MN$ denotes a matrix normal distribution and $\mbox{IW}$ and inverse Wishart.  Here, $M$ defines the mean matrix which $\Sigma$ and $K$ define the left and right covariances of dimensions $F\times F$ and $J\times J$, respectively.  Typically, $K$ is assumed to be diagonal with $K = \mbox{diag}(k_1,\dots,k_F)$.

Using the fact that our features are independently Gaussian distributed, we can write
\begin{align}
       \mathbf{Y} \mid X,\Sigma,\Phi \sim MN(X\Phi,\Sigma,I_J)
\end{align}
The prior for ${X,\Sigma}$ above is conjugate to this likelihood, so we can analytically compute the marginal likelihood. Standard matrix normal inverse Wishart conjugacy results yield
\begin{align}
       P(\mathbf{Y} \mid \Phi) &= \frac{|K|^{F/2}|S_0|^{n_0/2}2^{VF/2}}{(2\pi)^{V/2}|S_{\bar{y}\bar{y}}|^{F/2}|S_0 + S_{y|\bar{y}}|^{(n_0 +V)/2}}\nonumber\\
       &\hspace{0.1in}\cdot \prod_{\ell = 1}^F \frac{\Gamma\left(\frac{n_0+1 - \ell}{2} + \frac{V}{2}\right)}{\Gamma\left(\frac{n_0+1 - \ell}{2} + \frac{V}{2}\right)}.
       \label{eqn:Ymarglike}
\end{align}
Here,
\begin{align}
       S_{y|\bar{y}} &= S_{yy} - S_{y\bar{y}}S_{\bar{y}\bar{y}}^{-1}S_{y\bar{y}}'\\
       S_{\bar{y}\bar{y}} &= \Phi\Phi' + K\\
       S_{y\bar{y}} &= \mathbf{Y}\Phi' + MK\\
       S_{yy} &= \mathbf{Y}\mathbf{Y}' + MKM'.
\end{align}

In the case of an unbounded number of concepts, we restrict our attention to noise covariances $\Sigma$ that are diagonal ($\Sigma = \mbox{diag}(\sigma_1^2,\dots,\sigma_F^2)$).  This implies
\begin{align}
       X_{kj} \mid \sigma_k^2,k_j &\sim N(M_{kj},\sigma_k^2/k_j)\\
       Y_{ki} \mid X,\Phi,\sigma_k^2 &\sim N(X_{k\cdot}\Phi_i,\sigma_k^2),
\end{align}
independently for all $i,j,k$.

Although we are not working with a finite model, the $\Phi$ matrix implicitly truncates our model.  Let $X_C$ represent the set of latent concept features associated with the instantiated concepts and $\Phi_C$ the non-zero columns of $\Phi$ associated with the active concepts. Then, our model above is equivalent to
\begin{align}
       \mathbf{Y} | X_C,\Phi_C \sim MN(X_C\Phi_C, \Sigma, I_C).
\end{align}
Using the marginal likelihood formula of Eq.~\eqref{eqn:Ymarglike}, and simplifying based on the factorization of the likelihood across the dimensions of our feature space, yields
\begin{align}
       P(Y|\Phi) = &\frac{|K_{\matC\matC}|^{F/2} \beta_\sigma^{\alpha_\sigma
           F} \Gamma(\alpha_\sigma +
         V/2)^F}{|\Phi_{\matC\cdot}\Phi_{\matC\cdot}^T +
         K_{\matC\matC}|^{F/2} (2\pi)^{FV/2} \Gamma(\alpha_\sigma)^F} \prod_{k=1}^F \hat{\beta}_k^{-(\alpha_\sigma + V/2)}.
\end{align}

\subsection{Synthetic Example}
\begin{figure*}[ht]
\begin{center} 
\subfigure[Ground truth \label{fig:weakGT}]{\includegraphics[width=4.3cm]{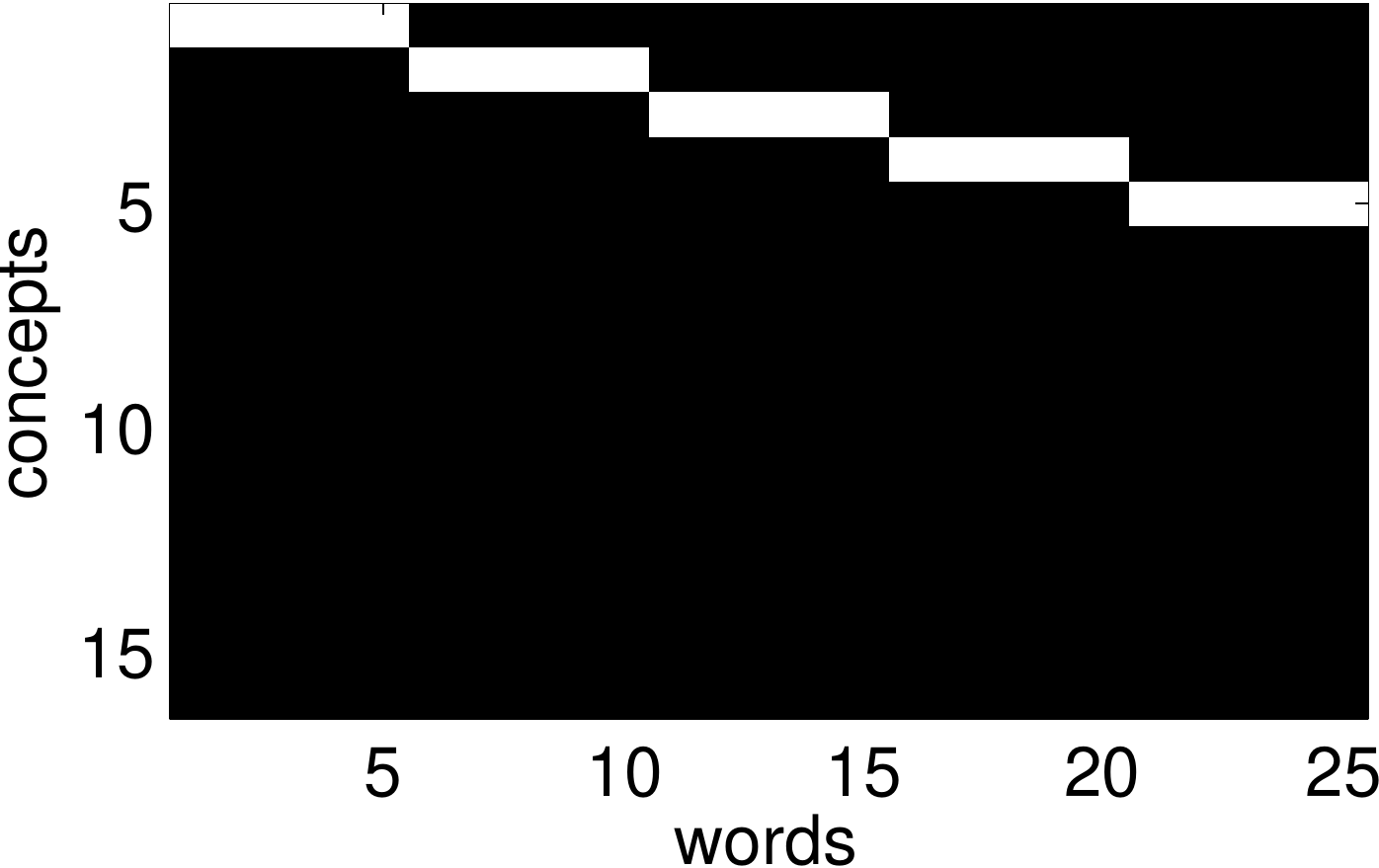}}
\subfigure[No semantic features \label{fig:weakNoPhi}]{\includegraphics[width=4.3cm]{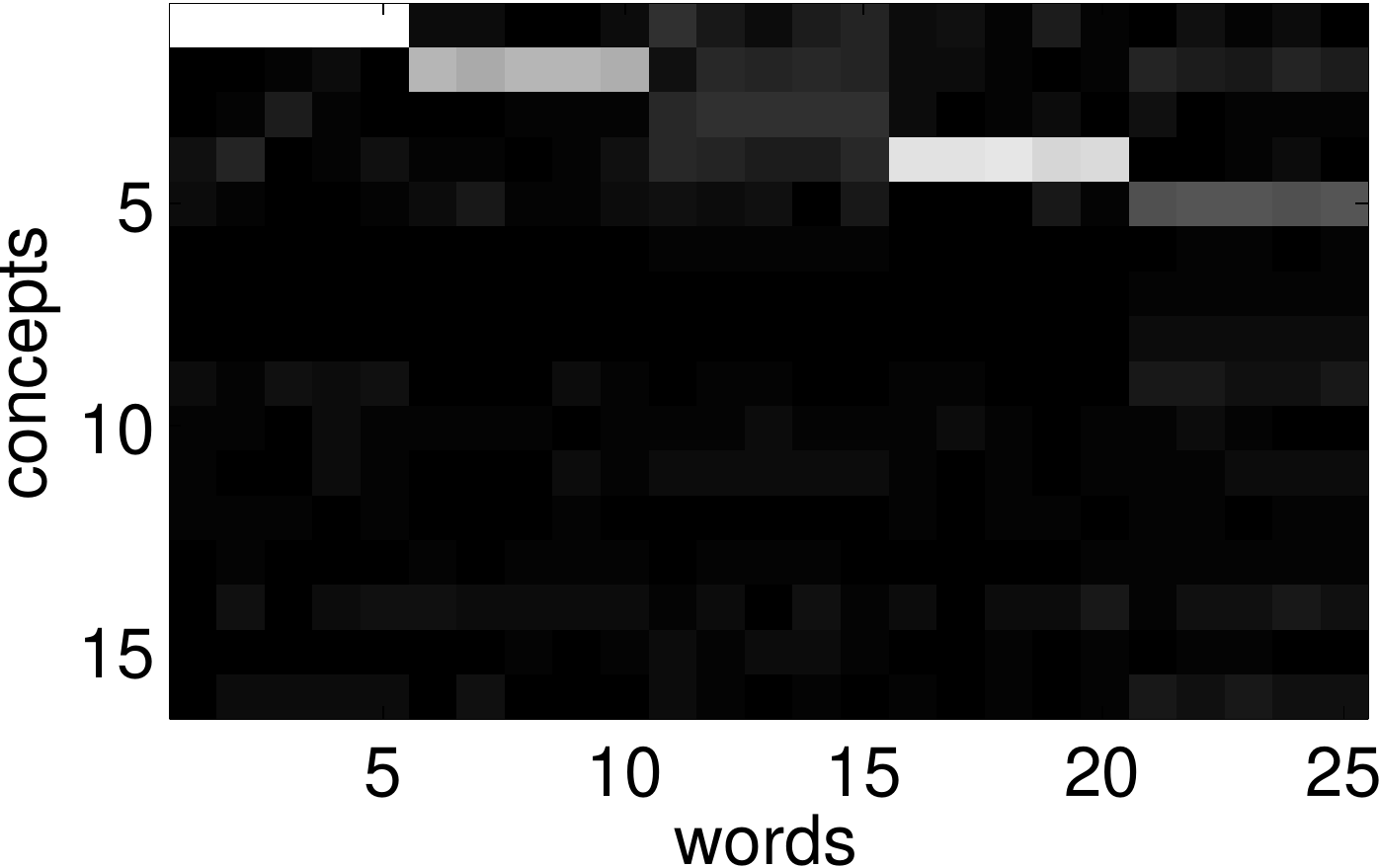}}
\subfigure[With semantic features \label{fig:weakPhi}]{\includegraphics[width=4.3cm]{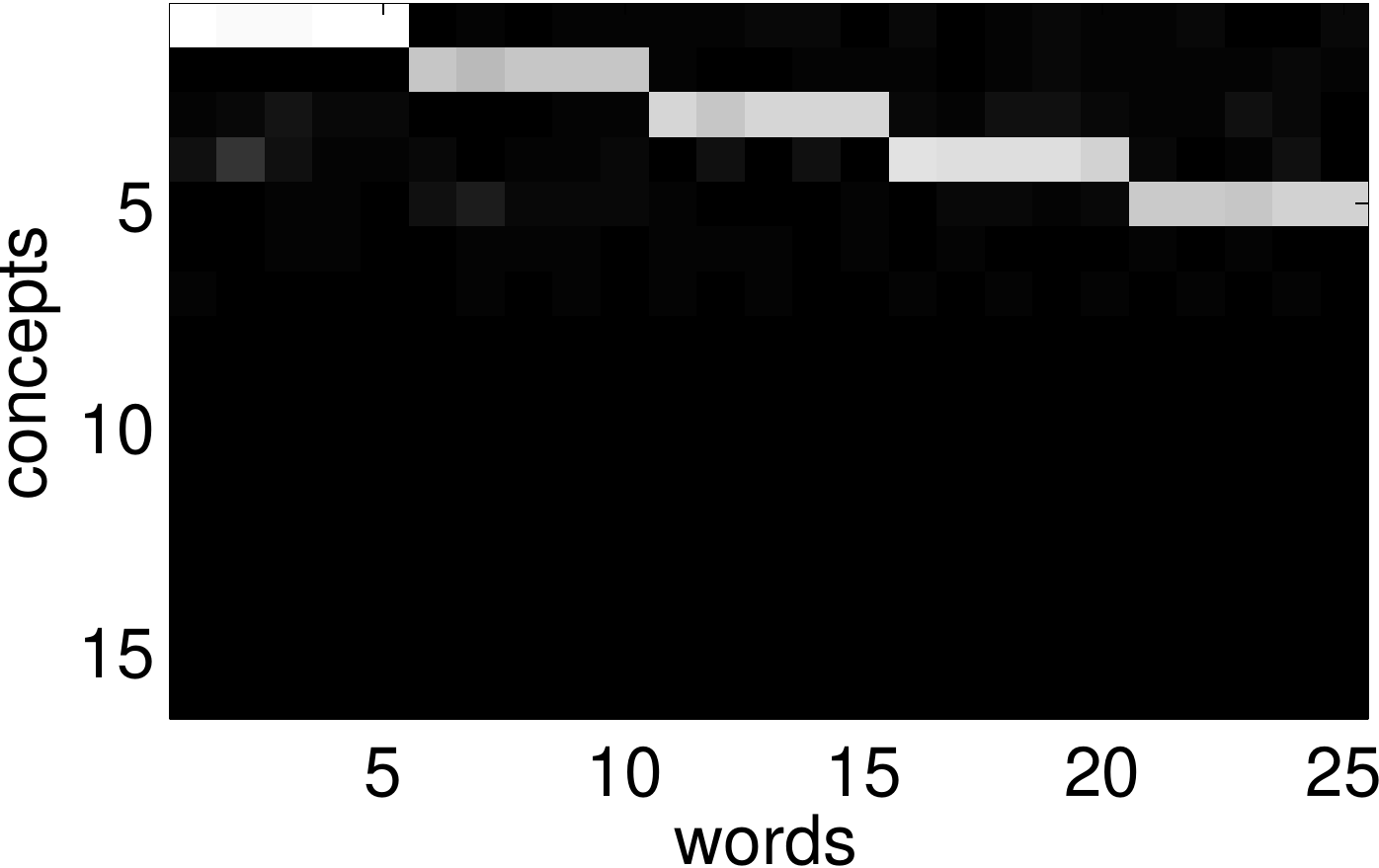}}
\vspace*{-5mm}
\caption{\footnotesize{Synthetic example illustrating weak identifiability of the
  nested beta process and how we overcome it using semantic
  features.  For each active concept, the average \fvD\ vector over
  all documents $d$ is plotted, with white indicating presence of a
  word in a concept, black indicating absence, and gray for average
  values of \fvD\ in  between 0 and 1.\label{fig:weakIdentify}}}
\vspace*{-5mm}
\end{center}
\end{figure*} 

The phenomenon of weak identifiability is illustrated in
Figure~\ref{fig:weakIdentify}.  We generate synthetic data for 100
documents, assuming the underlying concept representation depicted in
Figure~\ref{fig:weakGT}.  We then run our Gibbs sampler
(cf. Sec.~\ref{sec:sampling})  for 1,000
samples in order to infer the concept definitions for each document.
Figure~\ref{fig:weakNoPhi} shows the average concept definitions
\fvD\ across all documents, for the final sample.  We see
that while most of the concepts are correctly recovered, the sampler has difficulty reconstructing the
concept representing words 11 to 15.  However, by incorporating additional
semantic information about our vocabulary in the manner described
below, we are able to properly recover all five concepts, as seen in
Figure~\ref{fig:weakPhi}.

\subsection{User study results}
We filtered user study participants to make sure they had
followed the health care debate at the least at the level of reading
the headlines.  Of these 34 participants, we first measured
how many of the ten questions resulted in a favorable vote for our
model as compared to HDP, and found it to be 5.74 on average:
\begin{verbatim}
Num: 34
Average: 5.7352941176471
Standard Dev: 1.5398400132598
T-test 95% conf. interval:
  [5.2176965657521, 6.252891669542]
\end{verbatim}

We then asked, how many participants preferred our wordles to HDP,
treating each participant as a Bernoulli sample, and ignoring the
ties:
\begin{verbatim}
Num wins: 16
Num losses: 6
Binomial mean: 0.72727272727273
Binomial std: 0.44536177141512
Binomial Sign Test 95% conf. interval: 
  [0.54116788781464, 0.91337756673082]}
\end{verbatim}
\subsection{Sampler details}
\begin{figure*}[t]
\begin{center}
\subfigure[K-means initialization]{\includegraphics[width=4.3cm]{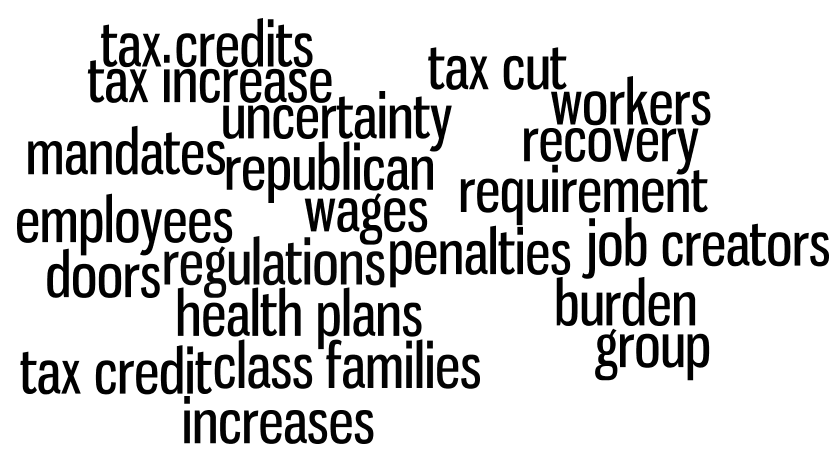}}
\subfigure[MAP sample]{\includegraphics[width=4.3cm]{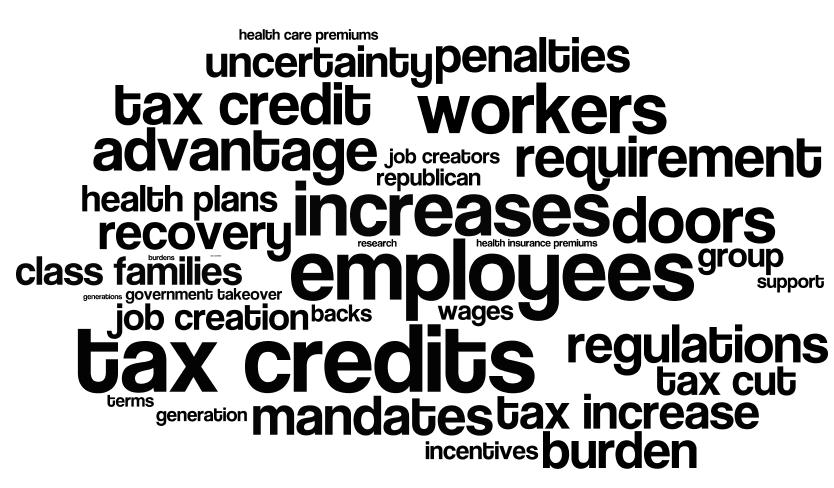}}
\vspace*{-5mm}
\caption{\footnotesize{Concept changes from initialization\label{fig:kmeans-comparison}}}
\vspace*{-5mm}
\end{center}
\end{figure*}

We initialize $\mathbf{\hat{c}}$ and $\mathbf{\hat{f}}$ in our sampler from a simple
k-means clustering of the words using the semantic features.  It is
interesting to see how, after many samples, how a concept's definition
changes from the initialization.  Figure~\ref{fig:kmeans-comparison}
shows one particular concept, and how our model reweights the words
and adds/removes words from the initial cluster.

We ran our sampler with the following hyperparameter settings:

\begin{verbatim}

\end{verbatim}



\section*{Supplementary material}

**In this user study, you will be presented with 10 pairs of word clouds describing various topics from the U.S. health care reform debate of 2009-2010, and will be asked a few questions. This study should take 5-10 minutes.**

### Do you follow US politics?

Yes   No

**How much did you pay attention to the US health care debate of 2009-2010?**

- I followed it closely.
- I paid attention to the main headlines.
- Very little, if at all.

**1. Which of the two word clouds more coherently describes the concept COSTS?**

this one

this one

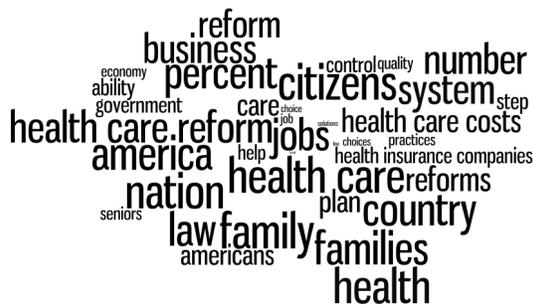
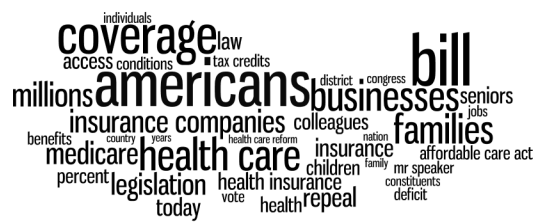

**2. Which of the two word clouds more coherently describes the concept REPEAL?**

[this one](#)

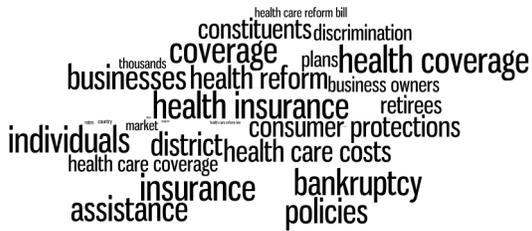

[this one](#)

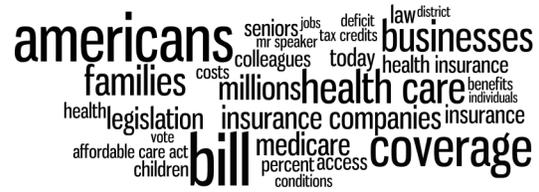

---

**3. Which of the two word clouds more coherently describes the concept PREMIUMS?**

[this one](#)

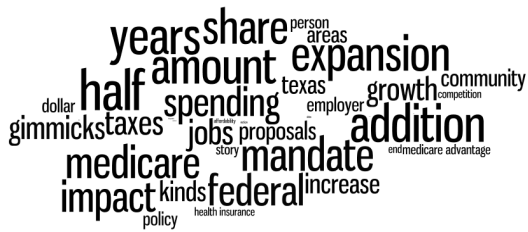

[this one](#)

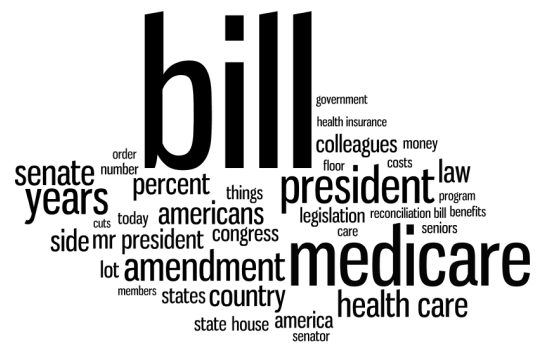

---

**4. Which of the two word clouds more coherently describes the concept MANDATES?**

[this one](#)

[this one](#)

[Word cloud 1: businesses, taxes, jobs, business, cbo, economy, percent, tax, workers, motion, reconciliation bill, employer mandate, employees, wages, health care reform, insurance, investment, etc.]

[Word cloud 2: tax credits, employees, increases, recovery, uncertainty, burden, tax increase, job creation, incentives, regulations, requirement, workers, tax credit, tax cut, families, penalties, health plans, etc.]

**5. Which of the two word clouds more coherently describes the concept INSURANCE?**

this one

[Word cloud: health coverage, health insurance, district, assistance, discrimination, health reform, health care coverage, retirees, plans, health care costs, businesses, business owners, consumer protections, thousands, constituents, individuals, bankruptcy, repeal, coverage, policies]

this one

[Word cloud: americans, coverage, health care, bill, health insurance, deficit, children, millions, law, affordable care act, jobs, vote, legislation, tax credits, conditions, today, benefits, costs, insurance companies, district, percent, families, access, seniors, health, businesses, medicare, colleagues, repeal]

**6. Which of the two word clouds more coherently describes the concept MEDICARE?**

this one

[Word cloud: bill, president, amendment, health care, percent, years, senate, americans, america, law, senator, program, reconciliation bill, state, floor, mr president, things, congress, premiums, costs, benefits, money, today, house, states, members, government, side, seniors, country, colleagues, number, order, lot, legislation, cuts, health insurance]

this one

[Word cloud: premiums, years, expansion, federal, spending, amount, impact, mandate, addition, share, increase, half, taxes, community, medicare advantage, employer, gimmicks, areas, competition, health insurance, policy, proposals, growth, story, end, texas, person, kinds, dollar, jobs]

**7. Which of the two word clouds more coherently describes the concept STATES?**

[this one]

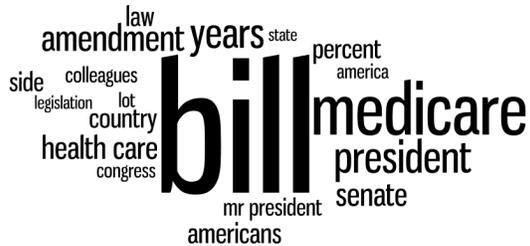

[this one]

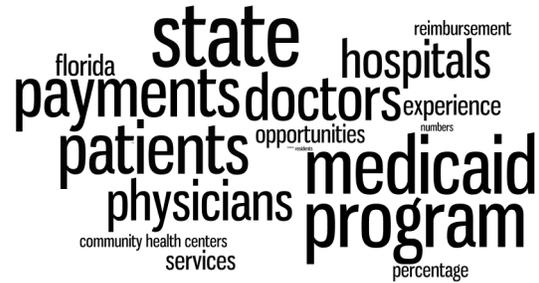

**8. Which of the two word clouds more coherently describes the concept EMPLOYERS?**

[this one]

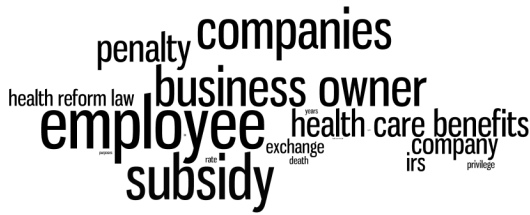

[this one]

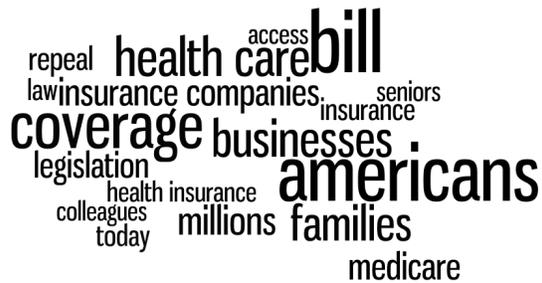

**9. Which of the two word clouds more coherently describes the concept OBAMACARE?**

[this one]   [this one]

*(word cloud 1: care, mr speaker, costs, doctors, government, plan, patients)*

*(word cloud 2: money, mistake, health care law, medicine, bill, challenge)*

---

**10. Which of the two word clouds more coherently describes the concept COVERAGE?**

[this one]   [this one]

*(word cloud 3: medicare, bill, health care, americans, individuals, colleagues, affordable care act, deficit, vote, conditions, repeal, percent, costs, mr speaker, health, children, families, insurance companies, seniors, access, legislation, businesses, benefits, millions, health insurance, jobs, insurance, law, tax credits, district, today)*

*(word cloud 4: policies, businesses, plans, health care coverage, insurance, health reform, retirees, discrimination, assistance, bankruptcy, repeal, market, health coverage, district, health care costs, consumer protections, thousands, health insurance, business owners, individuals, constituents, health care reform bill)*

---

**Make sure you've made a selection for each one!**

[Submit]


\end{document}